\documentclass{article}

\usepackage[preprint]{neurips_2026}

\usepackage[utf8]{inputenc}
\usepackage[T1]{fontenc}
\usepackage{hyperref}
\usepackage{url}
\usepackage{booktabs}
\usepackage{amsfonts}
\usepackage{amsmath}
\usepackage{nicefrac}
\usepackage{microtype}
\usepackage{xcolor}
\usepackage{graphicx}
\usepackage{tikz}
\usetikzlibrary{positioning, calc, arrows.meta}
\usepackage{multirow}
\usepackage{enumitem}
\usepackage{placeins}

\usepackage{soul}
\newif\ifshowcomments
\showcommentstrue
\ifshowcomments
  \newcommand{\cc}[1]{{\color{purple}[CC: #1]}}   
  \newcommand{\ccd}[1]{{\color{red}\st{#1}}}       
  \newcommand{\ug}[1]{{\color{blue}[UG: #1]}}      
\else
  \newcommand{\cc}[1]{}
  
  \newcommand{\ccd}[1]{}
  \newcommand{\ug}[1]{}
\fi

\title{The Bicameral Model: Bidirectional Hidden-State Coupling Between Parallel Language Models}

\author{%
  Cedric Flamant\thanks{Correspondence to: \texttt{cflamant@amazon.com}} \quad
  Udaya Ghai \quad
  Kanna Shimizu \\[0.5em]
  AWS Agentic AI
}

\begin{document}

\maketitle

\begin{abstract}
Existing multi-model and tool-augmented systems communicate by generating text, serializing every exchange through the output vocabulary. Can two pretrained language models instead coordinate through a continuous, concurrent channel? The \emph{Bicameral Model} couples two frozen language models through a trainable neural interface on their intermediate hidden states. At every generation step, both models run in lockstep: a primary model drives the task while an auxiliary model operates tools, solves constraints, or executes code, with both conditioning on each other's activations through a translation network and a learned suppression gate (${\sim}$1\% of combined parameters). The gate learns a selective communication protocol from task loss alone, without a prescribed format. We demonstrate the mechanism across three tool backends. On arithmetic, coupling two 0.5B models with a calculator raises accuracy from 36\% to 96\%. On logic grid puzzles, coupling two 0.6B models with a Z3 solver achieves $1.7\times$ the unaugmented baseline on ZebraLogic. On mathematical reasoning, coupling with a Python sandbox enables the auxiliary to generate problem-specific code from hidden-state signals alone, without ever seeing the problem text.
\end{abstract}

\section{Introduction}
\label{sec:intro}

Today, when two pretrained language models must cooperate, they communicate by generating text, serializing every exchange through discrete tokens. This paper asks whether two frozen LLMs can instead coordinate through their intermediate hidden states, and what the learned coupling looks like when trained only from task loss.

The Bicameral Model couples two language models through a trainable neural interface that provides a bridge for activations to flow between their intermediate hidden states. A primary model~$M_p$ orchestrates the task and produces the final response while an auxiliary model~$M_a$ maintains a concurrent stream with access to external tools. Both models advance in lockstep at every generation step, so the two streams condition on each other simultaneously and bidirectionally rather than through serialized text round-trips. The interface~$\phi$ sits between configurable transformer layers, translating activations between the models' representation spaces and modulating the coupling strength through a learned suppression gate. Both models' weights remain frozen; only $\phi$ is trained via dual-target supervised fine-tuning.

The primary model continues generating while the auxiliary operates tools in parallel, with results delivered through the neural channel rather than a text round-trip. We test the mechanism across three tool backends chosen to create a large, unambiguous capability gap between primary and auxiliary: a calculator, the Z3 constraint solver, and a Python sandbox. Tools are the device we use to produce this gap, not the mechanism itself; the same hidden-state coupling should function for any pair of complementary models. Coupling a 0.5B model with a calculator-equipped auxiliary raises arithmetic accuracy from 36.2\% to 96.5\%; coupling a 0.6B model with a Z3 solver achieves a $1.7\times$ improvement over the unaugmented model on ZebraLogic~\citep{lin2025zebralogic}; and in the Python-tool configuration, the auxiliary generates correct problem-specific code from hidden-state signals alone, without ever seeing the problem text.

Trained only on next-token loss, with no prescribed format for what the hidden-state channel should carry, the suppression gate converges to a selective, directionally structured communication protocol: forward coupling ($M_p \to M_a$) concentrates on task-relevant tokens such as numbers and operation keywords, while reverse coupling ($M_a \to M_p$) spikes when tool output becomes available or is recalled (§\ref{sec:analysis}). Moreover, this channel develops during training in a strict causal order, forward coupling first, then tool recall, then accuracy, consistent with the architecture's dependency structure. Adapter-equivalent ablations confirm the gains come from the auxiliary model's reasoning rather than from the trainable interface capacity.


\section{Architecture}
\label{sec:architecture}


\begin{figure}[t]
\centering
\definecolor{forcedgreen}{HTML}{d4edda}
\begin{tikzpicture}[
  box/.style={draw, rectangle, minimum width=2.5cm, minimum height=3cm, align=center, fill=cyan!20},
  toolbox/.style={draw, rounded corners, rectangle, minimum width=2.2cm, minimum height=1.5cm, align=center, fill=forcedgreen, font=\small},
  interface/.style={draw, rectangle, minimum width=1cm, minimum height=3cm, align=center, fill=red!20},
  arrow/.style={->, >=Stealth, thick},
  bluearrow/.style={->, >=Stealth, thick, color=blue!70!black},
  fatbluearrow/.style={->, >=Stealth, ultra thick, color=blue!70!black},
  purplearrow/.style={->, >=Stealth, thick, color=red!60!blue},
  fatpurplearrow/.style={->, >=Stealth, ultra thick, color=red!60!blue},
  blurplearrow/.style={->, >=Stealth, ultra thick, color=blue!60!red},
  redarrow/.style={->, >=Stealth, thick, color=red!70!black},
  lbl/.style={font=\scriptsize},
]

\node[toolbox] (tools) {Tools\\[-2pt]{\scriptsize(Z3, Calc, Python)}};
\node[box, right=1.8cm of tools] (aux) {};
\node[interface, right=1.8cm of aux] (neural) {\rotatebox{90}{\small Neural Interface $\phi$}};
\node[box, right=1.8cm of neural] (primary) {};

\node[above=0.55cm of aux, font=\small\bfseries] {Auxiliary $M_a$};
\node[above=0.55cm of primary, font=\small\bfseries] {Primary $M_p$};

\foreach \y in {0.6, 1.2, 1.8, 2.4} {
  \draw[gray!70] (aux.south west) ++(0,\y) -- ++(2.5,0);
}
\foreach \y in {0.6, 1.2, 1.8, 2.4} {
  \draw[gray!70] (primary.south west) ++(0,\y) -- ++(2.5,0);
}

\coordinate (aux-L2) at ([yshift=0.9cm]aux.south);
\coordinate (aux-L4) at ([yshift=2.1cm]aux.south);
\coordinate (pri-L2) at ([yshift=0.9cm]primary.south);
\coordinate (pri-L4) at ([yshift=2.1cm]primary.south);

\draw[redarrow] (aux.south) -- (aux-L2);
\draw[fatpurplearrow] (aux-L2) -- (aux-L4);
\draw[purplearrow] (aux-L4) -- (aux.north);

\draw[fatbluearrow] (primary.south) -- (pri-L2);
\draw[bluearrow] (pri-L2) -- (pri-L4);
\draw[blurplearrow] (pri-L4) -- (primary.north);

\node[above=0.1cm of aux, draw, inner sep=2pt, fill=white, font=\scriptsize\ttfamily] (aux_out) {aux output};
\node[below=0.1cm of aux, draw, inner sep=2pt, fill=white, font=\scriptsize\ttfamily] (aux_in) {aux input};
\node[above=0.1cm of primary, draw, inner sep=2pt, fill=white, font=\scriptsize\ttfamily] {output token};
\node[below=0.1cm of primary, draw, inner sep=2pt, fill=white, font=\scriptsize\ttfamily] {input token};

\node[right=0.1cm of primary, rotate=90, anchor=north, font=\scriptsize] {layers};

\draw[arrow] (aux_out.west) to[out=150, in=0] ([yshift=0.25cm]tools.east);
\draw[arrow] ([yshift=-0.25cm]tools.east) to[out=0, in=210] (aux_in.west);

\draw[fatbluearrow] (pri-L2) -- (neural.east |- pri-L2);
\draw[fatbluearrow] (neural.west |- aux-L2) -- (aux-L2);

\draw[fatpurplearrow] (aux-L4) -- (neural.west |- aux-L4);
\draw[fatpurplearrow] (neural.east |- pri-L4) -- (pri-L4);

\node[lbl, left=0.0cm of pri-L2, anchor=west] {read};
\node[lbl, left=0.0cm of aux-L2, anchor=east] {write};
\node[lbl, left=0.0cm of pri-L4, anchor=west] {write};
\node[lbl, left=0.0cm of aux-L4, anchor=east] {read};

\node[draw, rectangle, fill=white, align=left, anchor=north west, font=\scriptsize] at (current bounding box.north west) {
  \tikz\node[draw, fill=cyan!20, minimum width=0.3cm, minimum height=0.4cm] {}; Frozen weights \\
  \tikz\node[draw, fill=red!20, minimum width=0.3cm, minimum height=0.4cm] {}; Trainable (${\sim}$1\% params) \\
  \tikz{\draw[->, >=Stealth, thick, color=blue!70!black] (0,0) -- (0.25,0); \draw[->, >=Stealth, thick, color=red!60!blue] (0.25,0) -- (0.5,0);}; Activation flow
};

\end{tikzpicture}
\caption{%
\textbf{Bicameral architecture.}
A frozen primary model $M_p$ and a frozen auxiliary model $M_a$ run in parallel, coupled through a lightweight trainable neural interface $\phi$.
At each generation step, $\phi$ reads hidden states from $M_p$ at layer $\ell_r^p$ and injects a perturbation into $M_a$ at layer $\ell_w^a$ (forward coupling), then reads from $M_a$ at layer $\ell_r^a$ and injects into $M_p$ at layer $\ell_w^p$ (reverse coupling).
The auxiliary model's output tokens are routed to external tools (Z3 solver, calculator); tool results are forced back as input tokens.
}
\label{fig:system}
\end{figure}
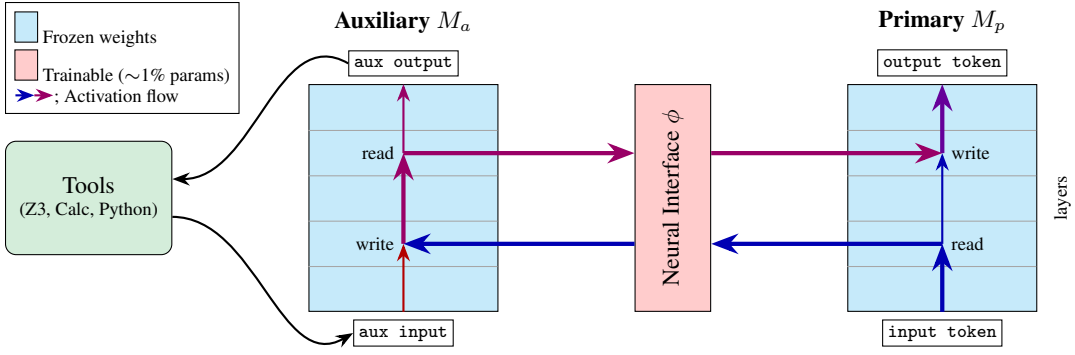

\paragraph{Notation.}
We use $m \in \{p, a\}$ to index the primary and auxiliary models. Each model is a pretrained transformer with $L_m$ blocks; we write its residual-stream state before block $\ell \in \{0, 1, \ldots, L_m\}$ executes as $\mathbf{h}_m^{(\ell)}(t) \in \mathbb{R}^{d_m}$, where $t$ is the token position. When $\ell = 0$ this is the input embedding, and when $\ell = L_m$ it is the state that feeds the final layer norm and unembedding; throughout the paper, ``layer~$\ell$'' refers to this residual-stream index. The tokens produced by each model at position $t$ are denoted $x_t^p$ and $x_t^a$ (the primary and auxiliary tokens; sources described in §\ref{sec:generation}). The interface reads and writes at four layer indices (read and write for each direction): $\ell_r^{p \to a}, \ell_w^{p \to a}, \ell_r^{a \to p}, \ell_w^{a \to p}$. Writing the forward direction for concreteness, the interface reads $\mathbf{h}_p^{(\ell_r^{p \to a})}(t)$ from $M_p$ and replaces $\mathbf{h}_a^{(\ell_w^{p \to a})}(t)$ in $M_a$ with a gated combination of the original state and a translated signal (§\ref{sec:neural_interface}); the reverse direction is symmetric. Our identity-coupling experiments constrain $\ell_r^{p \to a} = \ell_w^{p \to a}$ and $\ell_r^{a \to p} = \ell_w^{a \to p}$ (same-depth coupling, since identity requires $d_p = d_a$); \textsc{PullStandard} experiments sweep over general four-tuples.

\subsection{System overview}
\label{sec:system_overview}

The Bicameral Model consists of three components (Figure~\ref{fig:system}): a \emph{primary model}~$M_p$ that generates user-facing text, an \emph{auxiliary model}~$M_a$ that maintains a concurrent reasoning stream with access to external tools, and a \emph{neural interface} that couples the two through their intermediate hidden states. Both $M_p$ and $M_a$ are pretrained transformer language models whose weights are frozen; only the parameters of the interface are trained.

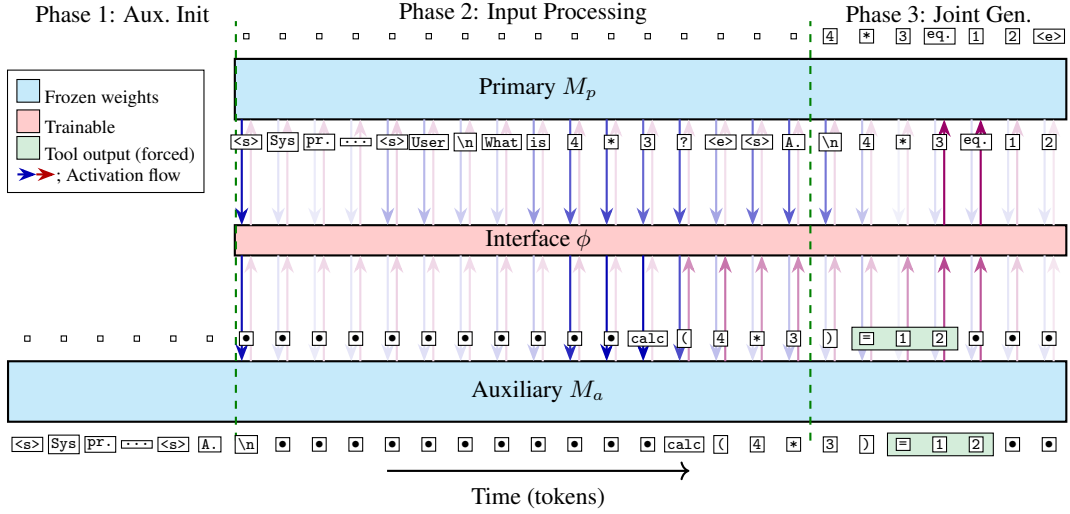
\begin{figure*}[t]
\centering
\definecolor{forcedgreen}{HTML}{d4edda}
\begin{tikzpicture}[
  bluearrow/.style={->, >=Stealth, thin, color=blue!70!black},
  fatbluearrow/.style={->, >=Stealth, ultra thick, color=blue!70!black},
  redarrow/.style={->, >=Stealth, thin, color=red!70!black},
  purplearrow/.style={->, >=Stealth, thin, color=red!60!blue},
  fatpurplearrow/.style={->, >=Stealth, ultra thick, color=red!60!blue},
  blurplearrow/.style={->, >=Stealth, ultra thick, color=blue!60!red},
]
  \def\topwidth{11}
  \def\bottomwidth{14}
  \def\normalheight{0.8}
  \def\midheight{0.4}
  \def\rightedge{0}
  \def\leftgreenline{-\topwidth}
  \def\rightgreenline{-3.4}
  \def\toppos{4}
  \def\midpos{2}
  \def\bottompos{0}
  \def\lineextend{0.5}
  \def\labelheight{5.0}
  \def\arrowgap{0.4}
  \def\arrowlength{4}
  \def\tokengap{0.3}
  \def\arrowoffset{0.06}

  \def\frozencolor{cyan!20}
  \def\hotcolor{red!20}
  \def\forcedcolor{forcedgreen}
  \def\tokenstyle{\tiny\ttfamily}

  \pgfmathsetmacro{\bottomleft}{\rightedge - \bottomwidth}
  \pgfmathsetmacro{\topleft}{\rightedge - \topwidth}
  \pgfmathsetmacro{\linetop}{\toppos + \normalheight/2 + \lineextend}
  \pgfmathsetmacro{\linebottom}{\bottompos - \normalheight/2 - \lineextend/2}
  \pgfmathsetmacro{\arinitcenter}{(\bottomleft + \leftgreenline)/2}
  \pgfmathsetmacro{\taskinitcenter}{(\leftgreenline + \rightgreenline)/2}
  \pgfmathsetmacro{\gencenter}{(\rightgreenline + \rightedge)/2}
  \pgfmathsetmacro{\arrowpos}{\linebottom - \arrowgap}
  \pgfmathsetmacro{\diagramcenter}{(\bottomleft + \rightedge)/2}
  \pgfmathsetmacro{\arrowstart}{\diagramcenter - \arrowlength/2}
  \pgfmathsetmacro{\arrowend}{\diagramcenter + \arrowlength/2}

  \pgfmathsetmacro{\taskoutputy}{\toppos + \normalheight/2 + \tokengap}
  \pgfmathsetmacro{\taskinputy}{\toppos - \normalheight/2 - \tokengap}
  \pgfmathsetmacro{\aroutputy}{\bottompos + \normalheight/2 + \tokengap}
  \pgfmathsetmacro{\arinputy}{\bottompos - \normalheight/2 - \tokengap}

  \node[draw, thick, minimum width=\topwidth cm, minimum height=\normalheight cm, anchor=east, fill=\frozencolor] (pri) at (\rightedge cm, \toppos cm) {};
  \node[font=\small] at (\diagramcenter cm, \toppos cm) {Primary $M_p$};

  \node[draw, thick, minimum width=\topwidth cm, minimum height=\midheight cm, anchor=east, fill=\hotcolor] (ni) at (\rightedge cm, \midpos cm) {};
  \node[font=\small] at (\diagramcenter cm, \midpos cm) {Interface $\phi$};

  \node[draw, thick, minimum width=\bottomwidth cm, minimum height=\normalheight cm, anchor=east, fill=\frozencolor] (aux) at (\rightedge cm, \bottompos cm) {};
  \node[font=\small] at (\diagramcenter cm, \bottompos cm) {Auxiliary $M_a$};

  \draw[green!50!black, thick, dashed] (\leftgreenline cm, \linebottom cm) -- (\leftgreenline cm, \linetop cm);
  \draw[green!50!black, thick, dashed] (\rightgreenline cm, \linebottom cm) -- (\rightgreenline cm, \linetop cm);

  \node[font=\small] at (\arinitcenter cm, \labelheight cm) {Phase 1: Aux.\ Init};
  \node[font=\small] at (\taskinitcenter cm, \labelheight cm) {Phase 2: Input Processing};
  \node[font=\small] at (\gencenter cm, \labelheight cm) {Phase 3: Joint Gen.};

  \def\imstart{\detokenize\expandafter{<s>}}
  \def\imend{\detokenize\expandafter{<e>}}
  \def\nl{\string\n}
  \def\arinputlist{\imstart, Sys, pr., \ldots, \imstart, A., \nl, \textbullet, \textbullet, \textbullet, \textbullet, \textbullet, \textbullet, \textbullet, \textbullet, \textbullet, \textbullet, \textbullet, calc, \string(, 4, \string*, 3, \string), =, 1, 2, \textbullet, \textbullet}
  \def\aroutputlist{ , , , , , , \textbullet, \textbullet, \textbullet, \textbullet, \textbullet, \textbullet, \textbullet, \textbullet, \textbullet, \textbullet, \textbullet, calc, \string(, 4, \string*, 3, \string), =, 1, 2, \textbullet, \textbullet, \textbullet}
  \def\taskinputlist{\imstart, Sys, pr., \ldots, \imstart, User, \nl, What, is, 4, \string*, 3, \string?, \imend, \imstart, A., \nl, 4, \string*, 3, eq., 1, 2}
  \def\taskoutputlist{ , , , , , , , , , , , , , , , , 4, \string*, 3, eq., 1, 2, \imend}

  \def\taskread{ 95, 11, 8, 10, 30, 28, 20, 16, 40, 64, 73, 60, 68, 35, 55, 64, 54, 23, 6, 13, 14, 10, 10}
  \def\taskwrite{ 15, 15, 15, 16, 18, 15, 14, 16, 13, 18, 14, 14, 15, 13, 14, 16, 16, 13, 12, 99, 100, 20, 10}
  \def\arread{ 15, 15, 15, 16, 18, 15, 14, 16, 13, 18, 14, 14, 44, 55, 43, 46, 26, 23, 42, 72, 67, 20, 10}
  \def\arwrite{ 95, 11, 8, 10, 11, 12, 12, 16, 22, 84, 100, 100, 70, 25, 15, 24, 14, 23, 6, 13, 14, 10, 10}

  \newcounter{artokencount}
  \foreach \token in \arinputlist { \stepcounter{artokencount} }
  \edef\numartokens{\theartokencount}

  \newcounter{tasktokencount}
  \foreach \token in \taskinputlist { \stepcounter{tasktokencount} }
  \edef\numtasktokens{\thetasktokencount}

  \pgfmathsetmacro{\tokenspacing}{\bottomwidth/\numartokens}

  \foreach \cval [count=\i] in \taskread {
    \draw[style={->, >=Stealth, thick, color=blue!70!black!\cval}] ({\rightedge - (\numtasktokens - \i + 0.5) * \tokenspacing - \arrowoffset}, \toppos - \normalheight/2) -- ({\rightedge - (\numtasktokens - \i + 0.5) * \tokenspacing - \arrowoffset}, \midpos + \midheight/2);
  }
  \foreach \cval [count=\i] in \taskwrite {
    \draw[style={->, >=Stealth, thick, color=red!60!blue!\cval}] ({\rightedge - (\numtasktokens - \i + 0.5) * \tokenspacing + \arrowoffset}, \midpos + \midheight/2) -- ({\rightedge - (\numtasktokens - \i + 0.5) * \tokenspacing + \arrowoffset}, \toppos - \normalheight/2);
  }
  \foreach \cval [count=\i] in \arwrite {
    \draw[style={->, >=Stealth, thick, color=blue!70!black!\cval}] ({\rightedge - (\numtasktokens - \i + 0.5) * \tokenspacing - \arrowoffset}, \midpos - \midheight/2) -- ({\rightedge - (\numtasktokens - \i + 0.5) * \tokenspacing - \arrowoffset}, \bottompos + \normalheight/2);
  }
  \foreach \cval [count=\i] in \arread {
    \draw[style={->, >=Stealth, thick, color=red!60!blue!\cval}] ({\rightedge - (\numtasktokens - \i + 0.5) * \tokenspacing + \arrowoffset}, \bottompos + \normalheight/2) -- ({\rightedge - (\numtasktokens - \i + 0.5) * \tokenspacing + \arrowoffset}, \midpos - \midheight/2);
  }

  \node[draw, fill=\forcedcolor, minimum width=1.4cm, minimum height=0.3cm] at (-2.15 cm, \aroutputy) {};
  \node[draw, fill=\forcedcolor, minimum width=1.4cm, minimum height=0.3cm] at (-1.68 cm, \arinputy) {};

  \foreach \token [count=\i] in \arinputlist {
    \node[draw, inner sep=1pt, fill=white, font=\tokenstyle] at ({\rightedge - (\numartokens - \i + 0.5) * \tokenspacing}, \arinputy cm) {\token};
  }
  \foreach \token [count=\i] in \aroutputlist {
    \node[draw, inner sep=1pt, fill=white, font=\tokenstyle] at ({\rightedge - (\numartokens - \i + 0.5) * \tokenspacing}, \aroutputy cm) {\token};
  }
  \foreach \token [count=\i] in \taskinputlist {
    \node[draw, inner sep=1pt, fill=white, font=\tokenstyle] at ({\rightedge - (\numtasktokens - \i + 0.5) * \tokenspacing}, \taskinputy cm) {\token};
  }
  \foreach \token [count=\i] in \taskoutputlist {
    \node[draw, inner sep=1pt, fill=white, font=\tokenstyle] at ({\rightedge - (\numtasktokens - \i + 0.5) * \tokenspacing}, \taskoutputy cm) {\token};
  }

  \draw[->, thick] (\arrowstart cm, \arrowpos cm) -- (\arrowend cm, \arrowpos cm);
  \node[below, font=\small] at (\diagramcenter cm, \arrowpos cm - 0.1 cm) {Time (tokens)};

  \node[draw, rectangle, fill=white, align=left, anchor=north west, yshift=-1.0cm, font=\scriptsize] at (current bounding box.north west) {
    \tikz\node[draw, fill=\frozencolor, minimum width=0.3cm, minimum height=0.3cm] {}; Frozen weights \\
    \tikz\node[draw, fill=\hotcolor, minimum width=0.3cm, minimum height=0.3cm] {}; Trainable \\
    \tikz\node[draw, fill=\forcedcolor, minimum width=0.3cm, minimum height=0.3cm] {}; Tool output (forced) \\
    \tikz{\draw[->, >=Stealth, thick, color=blue!70!black] (0,0) -- (0.25,0); \draw[->, >=Stealth, thick, color=red!70!black] (0.25,0) -- (0.5,0);}; Activation flow
  };

\end{tikzpicture}
\caption{%
\textbf{Unrolled generation.}
Token-by-token lockstep decoding for the arithmetic example ``What is $4 \times 3$?''
Arrow opacity reflects coupling strength (learned suppression gate $\sigma$).
\textbf{Phase~1:} $M_a$ processes its tool-instruction prompt (no coupling).
\textbf{Phase~2:} $M_p$ processes the user query; bidirectional coupling active (``wait'' tokens \textbullet\ on $M_a$).
\textbf{Phase~3:} Both models generate freely with bidirectional coupling.
$M_a$ emits \texttt{calc(4*3)}; the tool forces \texttt{=12} back as input tokens (green).
$M_p$ produces the answer ``4 * 3 equals 12'' without ever seeing calculator syntax.
}
\label{fig:unrolled}
\end{figure*}

The interface consists of two directional coupling operators, $\phi^{p \to a}: \mathbb{R}^{d_p} \times \mathbb{R}^{d_a} \to \mathbb{R}^{d_a}$ and $\phi^{a \to p}: \mathbb{R}^{d_a} \times \mathbb{R}^{d_p} \to \mathbb{R}^{d_p}$, each mapping a sender and receiver hidden state to an updated receiver state. Their concrete form (translation network plus learned suppression gate) is given in §\ref{sec:neural_interface}. A per-step forward pass then unfolds as (Figure~\ref{fig:unrolled}):
\begin{enumerate}[nosep,leftmargin=*]
\item Both models run from their input embeddings up to their forward-coupling layers, producing $\mathbf{h}_p^{(\ell_r^{p \to a})}(t)$ and $\mathbf{h}_a^{(\ell_w^{p \to a})}(t)$.
\item \emph{Forward coupling} updates $M_a$'s hidden state at $\ell_w^{p \to a}$:
\begin{equation}
\mathbf{h}_a^{(\ell_w^{p \to a})}(t) \;\gets\; \phi^{p \to a}\!\left(\mathbf{h}_p^{(\ell_r^{p \to a})}(t),\; \mathbf{h}_a^{(\ell_w^{p \to a})}(t)\right).
\label{eq:coupling_forward}
\end{equation}
\item Both models continue through the middle layers to the reverse-coupling layers, producing $\mathbf{h}_a^{(\ell_r^{a \to p})}(t)$ and $\mathbf{h}_p^{(\ell_w^{a \to p})}(t)$.
\item \emph{Reverse coupling} updates $M_p$'s hidden state at $\ell_w^{a \to p}$ by symmetry:
\begin{equation}
\mathbf{h}_p^{(\ell_w^{a \to p})}(t) \;\gets\; \phi^{a \to p}\!\left(\mathbf{h}_a^{(\ell_r^{a \to p})}(t),\; \mathbf{h}_p^{(\ell_w^{a \to p})}(t)\right).
\label{eq:coupling_reverse}
\end{equation}
\item Both models complete their remaining layers and produce a token (sampled or forced; see Section~\ref{sec:generation}).
\end{enumerate}
Forward coupling lets the auxiliary observe the primary's evolving representation; reverse coupling lets the auxiliary's reasoning flow back and influence primary generation. We write $\phi = \{\phi^{p \to a}, \phi^{a \to p}\}$ for the interface as a whole; $\phi$ is the only trainable component, and $M_p, M_a$ are frozen.

\subsection{Neural interface}
\label{sec:neural_interface}

We now unpack the coupling operators $\phi^{p \to a}$ and $\phi^{a \to p}$ introduced in §\ref{sec:system_overview}. Each direction has two trainable components: a \emph{translation network} $f^{\cdot}$ that maps a sender hidden state into the receiver's representation space, and a \emph{suppression gate} $g^{\cdot}$ that controls how much translated signal the receiver admits. Writing the forward direction for concreteness (reverse is symmetric):
\begin{align}
\phi^{p \to a}\!\left(\mathbf{h}_p, \mathbf{h}_a\right) \;=\; \big(1 - \sigma^{p \to a}\big) \, \mathbf{h}_a \;+\; \sigma^{p \to a} \, f^{p \to a}(\mathbf{h}_p), \qquad \sigma^{p \to a} \;=\; \mathrm{Sigmoid}\!\big(g^{p \to a}(\mathbf{h}_a)\big).
\label{eq:coupling_concrete}
\end{align}
Here $f^{p \to a}: \mathbb{R}^{d_p} \to \mathbb{R}^{d_a}$, and $g^{p \to a}: \mathbb{R}^{d_a} \to \mathbb{R}^{k}$ outputs a gate of dimension $k \in \{1, d_a\}$ (scalar or element-wise). The four networks $(f^{p \to a}, g^{p \to a}, f^{a \to p}, g^{a \to p})$ are the full parameterization of $\phi$.

\textbf{Pull design.} The gate reads the \emph{receiver's} hidden state, so the receiver decides how much external signal to admit based on its own state. Because Eq.~\ref{eq:coupling_concrete} is a convex combination with $\sigma \in [0, 1]$, the two extremes have interpretable meanings: at $\sigma \approx 0$ the receiver ignores the sender and proceeds as if uncoupled; at $\sigma \approx 1$ the receiver's hidden state is replaced entirely by the translated sender signal.\footnote{Preliminary experiments with ``push'' variants (gate computed from the sender) and gate-free variants also learn productive coupling; we report the pull design in what follows.}

\textbf{Instantiations.} We consider two concrete choices of $\phi$:
\begin{itemize}[nosep,leftmargin=*]
\item \textbf{\textsc{PullStandard}}: each $f^{\cdot}$ is a multi-layer perceptron and each $g^{\cdot}$ is a smaller MLP, both with ReLU activations. Both scalar and element-wise gates are used across our sweep. This is the primary variant used in our experiments.
\item \textbf{\textsc{PullIdentity}}: $f^{\cdot}$ is replaced by the identity map (requiring $d_p = d_a$) and only the suppression gates are trained. This variant requires models with compatible latent spaces, e.g.\ the same model coupled at matching depths.
\end{itemize}

\textbf{Layer adapters.}
In addition to $\phi$, some configurations insert low-rank bottleneck adapters~\citep{houlsby2019parameter} directly on the frozen models' own transformer layers. These adapters are distinct from $\phi$: each applies an additive residual $\mathbf{h} \gets \mathbf{h} + \frac{\alpha}{r} W_\mathrm{up}(\mathrm{ReLU}(W_\mathrm{down} \mathbf{h}))$ to a layer's output, with $W_\mathrm{up}$ zero-initialized so the adapter is a no-op at initialization. We use them only with identity-style interfaces, where they appear necessary for the coupled models to handle a parallel stream of superposed hidden states.


\subsection{Three-phase generation}
\label{sec:generation}

At inference time, both models advance in lockstep one token per position. At step $t$ each model emits one token: the primary token $x_t^p$ and the auxiliary token $x_t^a$. Each token comes from one of three sources:
\begin{itemize}[nosep,leftmargin=*]
\item \textbf{Sampled}, drawn from the model's conditional distribution with the coupling signals from $\phi$ already injected at the relevant hidden states (e.g.\ $x_t^p \sim M_p(\cdot \mid x_{<t}^p)$).
\item \textbf{Input-forced} from a given prefix (system prompt, user query). The forward pass is still run so the KV cache and coupling remain well-defined, but no sampling occurs at that position on that stream.
\item \textbf{Tool-forced} from an external tool (§\ref{sec:tools}) that has been invoked earlier in the auxiliary stream. These tokens are not sampled from $M_a$ and are masked from the auxiliary loss during training.
\end{itemize}
Bicameral generation proceeds in three phases (Figure~\ref{fig:unrolled}), which differ only in these source rules:
\begin{itemize}[nosep,leftmargin=*]
\item \textbf{Phase~1 (auxiliary prompt processing).} $x_t^a \gets$ prompt; the primary stream is inactive. Coupling disabled; this phase just populates $M_a$'s KV cache.
\item \textbf{Phase~2 (primary input processing).} $x_t^p \gets$ query; $x_t^a \sim M_a$ or $\gets$ tool output. Full bidirectional coupling is active so that both models build representations of the query before joint generation begins.
\item \textbf{Phase~3 (joint generation).} $x_t^p \sim M_p$; $x_t^a \sim M_a$ or $\gets$ tool output. Full bidirectional coupling is active. Generation terminates when the primary emits an end-of-sequence token.
\end{itemize}

\subsection{Tool integration}
\label{sec:tools}

The auxiliary model interacts with external tools during generation. We demonstrate three backends:

\textbf{Calculator.} The auxiliary model generates expressions in the format \texttt{calc(expr)}. The expression is evaluated and the result is forced back as tokens \texttt{=result;} into the auxiliary stream. For example, \texttt{calc(564*848)} produces \texttt{=478272;}.

\textbf{Z3 constraint solver.} We define ZebraDSL, a custom constraint language for logic grid puzzles, backed by the Z3 SMT solver~\citep{demoura2008z3}. The auxiliary model emits entity declarations, constraints, and queries; a pipeline (lexer, parser, semantic analyzer, Z3 translator) processes them and returns solver results. For example, \texttt{bob.color=red;} assigns an attribute and \texttt{?bob.house;} queries the solver (Appendix~\ref{app:zebradsl}).

\textbf{Python sandbox.} The auxiliary model generates \texttt{\textasciigrave\textasciigrave\textasciigrave python} code blocks. A subprocess-based sandbox executes the code and forces the output back as \texttt{\textasciigrave\textasciigrave\textasciigrave output} blocks. In all cases, tool output tokens are forced into the auxiliary stream and masked from the auxiliary loss during training. The primary model never sees tool output in its token stream; it receives influence only through the hidden-state coupling.

\section{Training}
\label{sec:training}

\subsection{Dual-target supervised fine-tuning}
\label{sec:dual_target_sft}

Both language models are frozen. Training optimizes only the neural interface parameters $\theta_\phi$. The loss is the sum of masked cross-entropy losses on both models' outputs:
\begin{equation}
\mathcal{L}(\theta_\phi) = \mathcal{L}_p(\theta_\phi) + \mathcal{L}_a(\theta_\phi)
\end{equation}
where $\mathcal{L}_p$ is computed over the primary model's response tokens and $\mathcal{L}_a$ is computed over the auxiliary model's generated content (forced tool outputs are masked).

Training uses teacher forcing on full sequences processed in parallel, just like regular SFT. The coupling architecture admits this because each model's forward pass decomposes around the coupling points: we run the pre-coupling layers of both models on their full input sequences in parallel, apply forward coupling to all hidden states at once, then run the post-coupling layers (including reverse coupling) up to the final token outputs.

\subsection{Causality-constrained data alignment}
\label{sec:causality}

The auxiliary model's actions must respect causality: it should not act on information the primary model has not yet produced at that token position. Similarly, if the primary model is trained to answer before it could have possibly received the tool output from the auxiliary model, it will never learn to use it. Training examples use a constraint solver to place auxiliary content (tool calls, reasoning steps) at valid positions relative to the primary token sequence. Causality tags mark the earliest and latest valid placement for each auxiliary content block, and a scheduling strategy (eager, lazy, random, or balanced) selects the final position within the valid window (Appendix~\ref{app:causality}).

\subsection{Reinforcement learning}
\label{sec:rl}

The architecture also supports reinforcement learning as a second phase after SFT. We implement GRPO~\citep{shao2024deepseekmath} over the interface parameters only (both LLMs remain frozen), using outcome-based reward with trajectories generated through the full coupled pipeline. Preliminary experiments on arithmetic show consistent improvements on held-out word problems (+4.7pp on GSM8K-IRL) with no regression on calculator-based arithmetic (Appendix~\ref{app:rl}).

\section{Experiments}
\label{sec:experiments}

We select domains where the tool creates an unambiguous capability gap, making the coupling's contribution easy to isolate from the base model's innate abilities. We evaluate the Bicameral Model in two domains: arithmetic with a calculator tool (Section~\ref{sec:arithmetic}) and logic grid puzzles with a Z3 constraint solver (Section~\ref{sec:zebra}).

\subsection{Arithmetic with calculator}
\label{sec:arithmetic}

We couple two frozen copies of Qwen2.5-0.5B-Instruct~\citep{yang2024qwen25} through a \textsc{PullStandard} interface, with only the auxiliary model having access to a calculator. We evaluate on general arithmetic, GSM8K~\citep{cobbe2021gsm8k}, and GSM8K-IRL, a harder rephrasing of GSM8K we generated with professionally-themed contexts and decimal quantities. Full experimental details are in Appendix~\ref{app:exp_details}.

\textbf{Main results.}
We sweep 890 sampled configurations, varying which coupling layers are selected. Table~\ref{tab:arithmetic} reports the best configuration alongside baselines. Of the 890 configurations, 95.4\% beat the primary-only arithmetic baseline, indicating robustness to layer choice. On the harder GSM8K-IRL benchmark, 23.7\% of configurations improve over the primary-only baseline (12.8\%), with the best reaching 17.1\% (+4.3pp).\footnote{The sweep includes configurations expected to underperform, such as shallow coupling and wirings without a within-token return path, to map the full design space; see Appendix~\ref{app:layer_sweep}.} On standard GSM8K, no configuration matches the primary-only baseline (49.6\%), likely because hidden-state perturbations disrupt reasoning patterns the base model already executes well and the calculator adds little value on simple arithmetic.

\textbf{Adapter-equivalent ablation.}
To test whether the gains come from the auxiliary model's calculator or merely from the additional 16M trainable parameters, we train an adapter-equivalent configuration (Appendix~\ref{app:adapter_equiv}): the same interface and training procedure, but with $M_a$ bypassed so signals round-trip through $\phi$ alone ($M_p \to \phi \to M_p$). The adapter-equivalent scores 48.0\% on arithmetic (versus 96.5\% bicameral) and 9.2\% on GSM8K-IRL (versus 17.1\%), confirming that the auxiliary model's tool use, not the parameter count, drives the gains.

\begin{table}[t]
\centering
\caption{Arithmetic results (Qwen2.5-0.5B-Instruct). \textsc{PullStandard} bicameral row reports the best of 890 layer-wiring configurations; 95.4\% beat the primary-only arithmetic baseline. Identity variants use same-depth coupling at layers 10/15. The adapter-equivalent bypasses the auxiliary model, feeding the forward network's output directly to the reverse network.}
\label{tab:arithmetic}
\small
\begin{tabular}{lrccc}
\toprule
Configuration & Params & Arith. & GSM8K & GSM8K-IRL \\
\midrule
Primary-only (no interface) & --- & 36.2 & \textbf{49.6} & 12.8 \\
\midrule
\multicolumn{5}{l}{\textit{Learned translation (\textsc{PullStandard})}} \\
Adapter-equivalent & 16M & 48.0 & 40.9 & 9.2 \\
Bicameral (best of 890) & 16M & \textbf{96.5} & 39.1 & 17.1 \\
\midrule
\multicolumn{5}{l}{\textit{Identity coupling (same-depth, layers 10/15)}} \\
\textsc{ElemIdentity} + adapters & 919K & 84.2 & 40.8 & \textbf{22.2} \\
\textsc{ScalarIdentity} + adapters$^\dagger$ & 803K & 68.1{\small$\pm$11.8} & 36.8{\small$\pm$1.5} & 17.5{\small$\pm$3.0} \\
\textsc{ScalarIdentity}, no adapters & 115K & 39.0 & 42.8 & 11.1 \\
\bottomrule
\end{tabular}

\vspace{0.3em}
{\footnotesize $^\dagger$Mean $\pm$ std over 5 seeds. Other rows are single runs.}
\label{tab:architecture}
\end{table}

\textbf{Identity coupling.}
An alternative approach is \textsc{PullIdentity}: hidden states pass directly between models with only a suppression gate and low-rank adapters. With ${\sim}$800K--900K trainable parameters, identity interfaces sacrifice peak arithmetic accuracy (68--84\% vs 96.5\%) but \emph{outperform} learned translation on word problems, reaching 22.2\% on GSM8K-IRL (9.4pp above primary-only, 5.1pp above \textsc{PullStandard}). Operating directly in the models' native representation space without compressing through a learned bottleneck may be what helps generalization to out-of-distribution problems.

\subsection{Logic puzzles with Z3 solver}
\label{sec:zebra}

We couple two frozen copies of Qwen3-0.6B through a \textsc{PullStandard} interface. The auxiliary emits ZebraDSL, a small declarative language we designed for logic grid puzzles (e.g., discrete-choice assignment puzzles such as Sudoku or clue-based logic grids). ZebraDSL compiles to assertions for the Z3 SMT solver~\citep{demoura2008z3}, which checks whether the constraints are consistent and returns any implied values. We evaluate on ZebraLogic~\citep{lin2025zebralogic}, a benchmark of 1{,}000 logic grid puzzles, and on GeneralZebra, a dynamically generated in-distribution test suite. Full details and the complete ZebraDSL specification are in Appendices~\ref{app:exp_details} and~\ref{app:zebradsl}.

\textbf{Main results.}
Table~\ref{tab:zebralogic} presents four training configurations that vary in clue presentation style. Our best configuration (40 epochs, ZL-style clues) reaches $64.7 \pm 6.7$\% on ZebraLogic (mean $\pm$ std over 5 seeds), a $1.7\times$ improvement over the unaugmented Qwen3-0.6B baseline (37.5\% with thinking, 32k max tokens). For context, the ZebraLogic leaderboard reports 36.2\% for Claude~3.5 Sonnet and 31.7\% for GPT-4o, though these models lack solver access. Training without ZebraLogic-style clue presentation drops ZebraLogic accuracy to 10.3\% despite 81.1\% on GeneralZebra, indicating the auxiliary leans heavily on memorized clue patterns; the text-level tool baseline below suggests this rigidity reflects a Qwen3-0.6B capability ceiling rather than the architecture, since the model cannot use ZebraDSL as text at all without training. As a preliminary probe of entity generalization, a single attempt at replacing ZebraLogic entities with values absent from all training categories produced a correct solution (Figure~\ref{fig:full_zebra}); systematic OOD entity evaluation is left to future work.

\begin{table}[t]
\centering
\caption{ZebraLogic and GeneralZebra results. Primary and auxiliary models are both Qwen3-0.6B~\citep{yang2025qwen3}. All bicameral configurations use \textsc{PullStandard} with 21M interface parameters. Primary-only baselines evaluated at 32k tokens. Adapter-equivalent uses the same interface but bypasses the auxiliary model.}
\label{tab:zebralogic}
\small
\begin{tabular}{lcc}
\toprule
Configuration & ZebraLogic & GeneralZebra \\
\midrule
Primary-only (thinking, 32k) & 37.5 & 15.2 \\
Primary-only (no thinking, 32k) & 4.9 & 2.9 \\
Adapter-equivalent (40 ep.) & 7.5 & 7.9 \\
\midrule
Bicameral, w/o ZL clue style (20 ep.) & 10.3 & 81.1 \\
Bicameral, w/ ZL clue style (20 ep.) & 48.3 & 87.0 \\
Bicameral, ZL clue style only (20 ep.) & 45.3 & 8.0 \\
Bicameral, w/ ZL clue style (40 ep.) & \textbf{64.7 $\pm$ 6.7} & \textbf{93.2 $\pm$ 0.9} \\
\bottomrule
\end{tabular}
\end{table}

\begin{figure}[t]
\centering
\includegraphics[width=\linewidth]{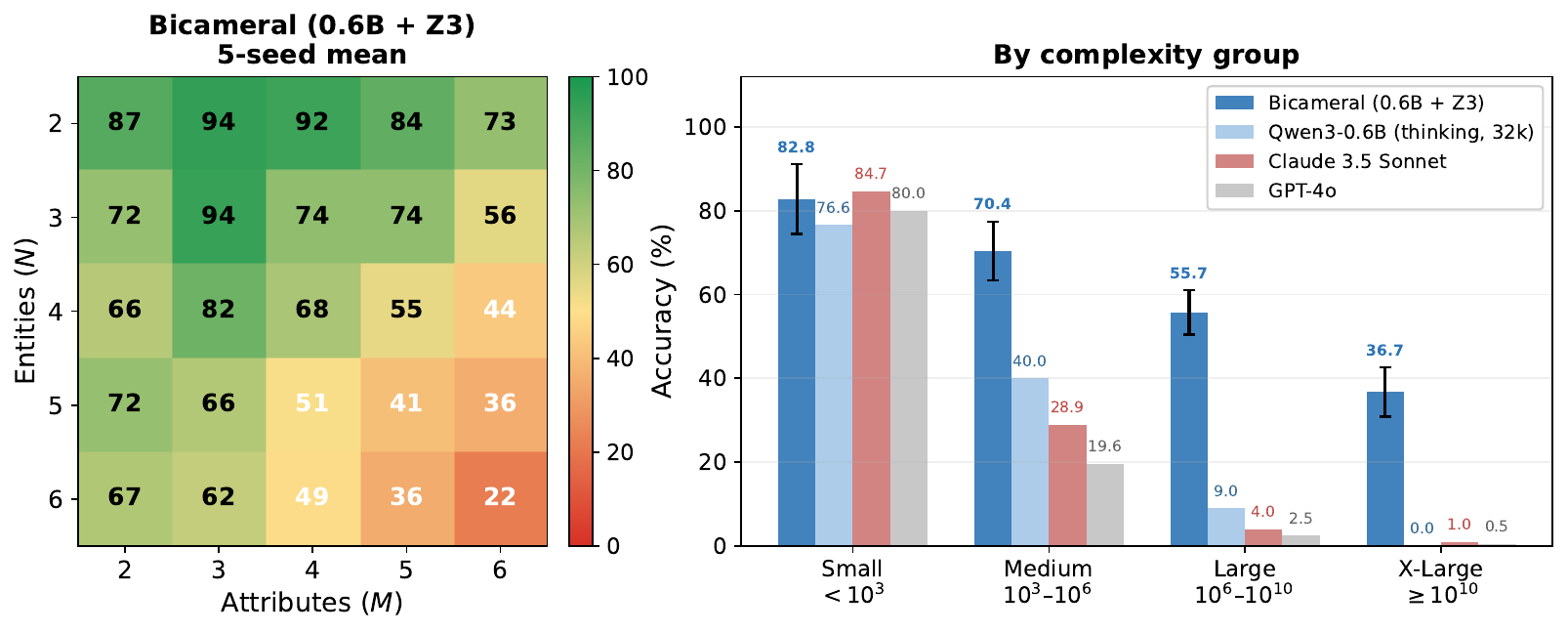}
\caption{ZebraLogic accuracy by puzzle complexity (5-seed mean). \textbf{Left:} Per-size heatmap ($n=40$ per cell per seed). \textbf{Right:} Grouped accuracy with seed std error bars; Claude~3.5 Sonnet and GPT-4o baselines from~\citet{lin2025zebralogic} shown for calibration (these models lack solver access).}
\label{fig:zebralogic_by_size}
\end{figure}

\textbf{Accuracy by puzzle size.}
Figure~\ref{fig:zebralogic_by_size} reports accuracy per puzzle size, grouped by ZebraLogic's search-space bins~\citep{lin2025zebralogic}. The bicameral system reaches $82.8 \pm 8.3$\% on small puzzles (search space $<10^3$) and drops to $36.7 \pm 5.8$\% on extra-large puzzles (search space $\geq 10^{10}$); the Claude~3.5 Sonnet and GPT-4o curves are plotted alongside for calibration, with the gap widening sharply as search space grows.

\textbf{Text-level tool baseline.}
We prompt thinking-enabled Qwen3-0.6B with ZebraDSL syntax instructions and pipe its text output through the ZebraDSL execution pipeline. This baseline scores 24.6\% (below the 37.5\% primary-only baseline) with zero successful Z3 queries across all 1{,}000 puzzles: all correct answers come from chain-of-thought, not the solver. Tool instructions alone do not enable Z3 use at this model size, though dedicated SFT would likely close this gap.

\section{Analysis and discussion}
\label{sec:analysis}

\begin{figure}[t]
\centering
\includegraphics[width=\linewidth]{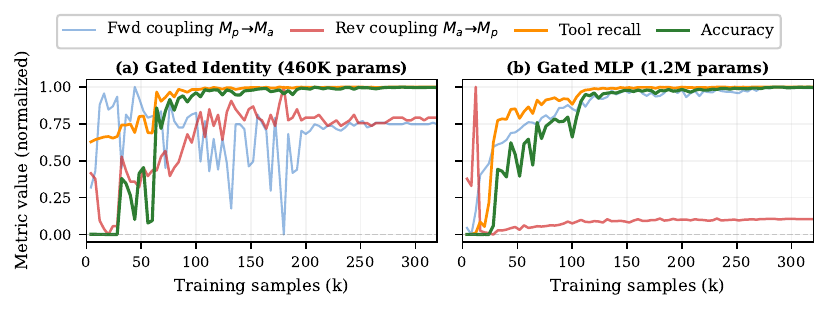}
\caption{%
\textbf{Communication onset during training.}
Normalized coupling strength, tool recall, and accuracy over 320k training samples (multiplication-only, $[1, 10^7]$, primary-only baseline 0\%) for (a)~Gated Identity (\textsc{ScalarIdentity}, 460K params) and (b)~Gated MLP (\textsc{PullStandard}, 1.2M params).
Forward coupling activates immediately; tool recall develops next; accuracy undergoes sudden onset once tool recall crosses a threshold.
Both architectures reach 99.9\% accuracy.
}
\label{fig:phase_transition}
\end{figure}

\textbf{Coordination develops as a phase transition.}
To understand \emph{how} the coupling develops, we train on multiplication-only data with inputs uniformly sampled over the range $[1, 10^7]$ and track four metrics: forward perturbation norm, reverse perturbation norm, tool recall (fraction of problems where the auxiliary calls the calculator with correct operands), and task accuracy (Figure~\ref{fig:phase_transition}). Two scalar-gated architectures are compared: the gated identity interface and the gated MLP interface. Both scalar gates output initially near zero, so coupling strength starts negligible and must be actively learned.

Both architectures show the same qualitative pattern. Forward coupling activates immediately (within 4k samples), establishing a channel from primary to auxiliary. Tool recall ramps next: the gated identity starts at 63\% (compatible raw hidden states already carry useful signal) and climbs to 97\% by 64k samples; the gated MLP starts at 0\% (the translation must be learned from scratch) and reaches 85\% by 32k samples. Accuracy undergoes a sudden onset once tool recall crosses a threshold: the system spends 28k samples at 0\% then jumps to 40--60\% within 4--8k samples, stabilizing above 95\% by 100--120k samples. Both architectures reach 99.9\%.

\begin{figure*}[t]
\centering
\includegraphics[width=\linewidth]{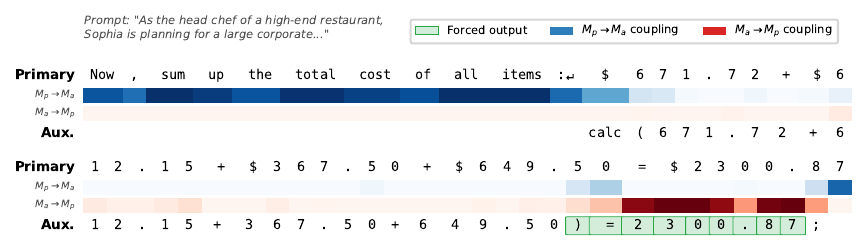}
\caption{%
\textbf{Coupling activity during arithmetic generation.}
Token-by-token coupling strength for a multi-step word problem (Qwen2.5-0.5B, scalar identity interface, tokens 464--515).
\textbf{Top two rows:} Primary model tokens with forward (blue) and reverse (red) coupling strength.
\textbf{Bottom two rows:} Auxiliary model tokens with the same coupling channels.
Forward coupling activates on relevant primary tokens (``sum,'' ``total,'' ``all items'') but not on the dollar amounts themselves, because the auxiliary has already received those values earlier in the sequence.
Reverse coupling is near zero until the calculator returns its result (\texttt{=2300.87}), at which point it spikes as the answer flows back to the primary model.
See Figure~\ref{fig:full_arithmetic} for the full trace.
}
\label{fig:coupling_activity}
\end{figure*}

\textbf{Selective coupling.}
The suppression gate learns a structured activation pattern from task loss alone, with no prescribed protocol. Figure~\ref{fig:coupling_activity} shows the coupling strength at each token position during a multi-step arithmetic problem. Forward coupling ($M_p \to M_a$, blue) activates on relevant tokens: ``sum,'' ``total,'' ``cost,'' and ``all items''. Notably, the forward channel does not activate on the dollar amounts themselves because the auxiliary already received those values earlier in the sequence and does not need them repeated. Reverse coupling ($M_a \to M_p$, red) is near zero throughout this period and spikes only when the calculator returns its result (\texttt{=2300.87}).

\textbf{Interface capacity trades in-distribution fit for OOD generalization.}
\textsc{PullStandard}'s learned translation achieves the highest in-distribution accuracy (96.5\% arithmetic) but is outperformed by identity coupling on out-of-distribution GSM8K-IRL (22.2\% vs 17.1\%, Section~\ref{sec:arithmetic}). The pattern is consistent with a capacity--generalization tradeoff: a freely learned translation fits the training distribution's exact representational structure but transfers less reliably than a near-identity pass-through with a learned gate, which leans on the inductive bias that two copies of the same pretrained model have compatible representations at matching depths.

\textbf{Mathematical reasoning with Python sandbox.}
Coupling twin Qwen3-4B models with a Python sandbox on NuminaMath-TIR~\citep{numina2024numinamath} yields 62.5\% on MATH~\citep{hendrycks2021math}, below the unaugmented Qwen3-4B-with-thinking baseline (81.6\% at 32k tokens). However, the coupling produces correct answers on 127 problems where thinking alone fails, and qualitative inspection of the hidden-state channel shows problem-specific code being generated (Appendices~\ref{app:python_examples},~\ref{app:full_visualizations}).

\textbf{Limitations.}
Lockstep generation can double compute requirements. The coupling degrades performance on tasks the base model already handles well: GSM8K drops from 49.6\% to around 40\% across all 890 configurations, consistent with the framing that when the capability gap is small, hidden-state perturbations inject more noise than signal. Training also requires task-specific data with causality annotations (§\ref{sec:conclusion} discusses a possible RL relaxation).

\section{Related work}
\label{sec:related}

\textbf{Latent-space communication between LLMs.}
A growing body of work replaces text-based inter-model communication with hidden-state transfer. \citet{ramesh2025activations} show that even a zero-parameter one-shot activation graft between frozen models improves over natural-language debate, but their coupling is unidirectional and applied once rather than at every generation step. C2C~\citep{fu2026c2c} fuses one model's KV-cache into another's via learned projections with per-layer Gumbel-sigmoid gating, again unidirectional and applied once at prefill. Interlat~\citep{du2025interlat} trains a communication adapter for latent transfer from a frozen sender to a fine-tuned receiver in a turn-based setup. LatentMAS~\citep{zou2025latentmas} transfers full KV caches between agents in a training-free sequential pipeline. CALM~\citep{bansal2024llmaugmented} composes two frozen models via cross-attention on intermediate representations. Concurrent work by \citet{yang2026recursive} replaces text-based multi-agent communication with latent-state transfer through trainable projections between frozen LLMs in a recursive pipeline, but their agents communicate in sequential discrete rounds. The Bicameral Model differs from all of these in providing \emph{bidirectional continuous coupling} at every decoding step.

\textbf{Multi-agent systems.}
Debate~\citep{du2024multiagent}, self-consistency~\citep{wang2023selfconsistency}, and mixture-of-agents communicate through generated text. ThoughtComm~\citep{zheng2025thoughtcomm} enhances multi-agent debate by injecting latent thoughts from agents' hidden states as prefix embeddings between rounds. These approaches are turn-based; bicameral coupling instead operates at every token step.

\textbf{Representation alignment.}
The Platonic Representation Hypothesis~\citep{huh2024platonic} argues that model representations converge across architectures, supporting the feasibility of learned projections between LLM latent spaces. FuseChat~\citep{wan2024fusechat} and Co-LLM~\citep{shen2024collm} compose models at the output level, not via intermediate hidden states.

\section{Conclusion}
\label{sec:conclusion}

Two frozen language models connected by a lightweight neural interface on their hidden states learn to cooperate on tasks neither can solve alone, across arithmetic, logic puzzles, and Python-assisted math. The communication is selective: the learned suppression gate fires on task-relevant tokens and stays silent otherwise, producing a structured protocol without a prescribed format; adapter-equivalent ablations confirm the gains come from the auxiliary's reasoning, not the interface parameters. Hidden-state coupling opens a design space for multi-model systems that communicate below the token level: the architecture should extend to other complementary-model pairings such as retrieval or non-text-modality models, and, with adaptations to the training objective, to reward or process-reward models delivering dense quality signal. A second direction is leaning further into RL on the interface (§\ref{sec:rl}): scaling to more domains and larger models, and relaxing SFT's causality-annotation requirements as RL picks up the coordination timing.


\bibliographystyle{plainnat}
\bibliography{references}

@inproceedings{lin2025zebralogic,
  title={ZebraLogic: On the Scaling Limits of {LLMs} for Logical Reasoning},
  author={Lin, Bill Yuchen and Le Bras, Ronan and Richardson, Kyle and Sabharwal, Ashish and Poovendran, Radha and Clark, Peter and Choi, Yejin},
  booktitle={International Conference on Machine Learning (ICML)},
  year={2025}
}

@inproceedings{schick2023toolformer,
  title={Toolformer: Language Models Can Teach Themselves to Use Tools},
  author={Schick, Timo and Dwivedi-Yu, Jane and Dess{\`i}, Roberto and Raileanu, Roberta and Lomeli, Maria and Hambro, Eric and Zettlemoyer, Luke and Cancedda, Nicola and Scialom, Thomas},
  booktitle={Advances in Neural Information Processing Systems (NeurIPS)},
  year={2023}
}

@inproceedings{gou2024tora,
  title={{ToRA}: A Tool-Integrated Reasoning Agent for Mathematical Problem Solving},
  author={Gou, Zhibin and Shao, Zhihong and Gong, Yeyun and Shen, Yelong and Yang, Yujiu and Huang, Minlie and Duan, Nan and Chen, Weizhu},
  booktitle={International Conference on Learning Representations (ICLR)},
  year={2024}
}

@article{cobbe2021gsm8k,
  title={Training Verifiers to Solve Math Word Problems},
  author={Cobbe, Karl and Kosaraju, Vineet and Bavarian, Mohammad and Chen, Mark and Jun, Heewoo and Kaiser, Lukasz and Plappert, Matthias and Tworek, Jerry and Hilton, Jacob and Nakano, Reiichiro and Hesse, Christopher and Schulman, John},
  journal={arXiv preprint arXiv:2110.14168},
  year={2021}
}

@inproceedings{hendrycks2021math,
  title={Measuring Mathematical Problem Solving with the {MATH} Dataset},
  author={Hendrycks, Dan and Burns, Collin and Kadavath, Saurav and Arora, Akul and Basart, Steven and Tang, Eric and Song, Dawn and Steinhardt, Jacob},
  booktitle={Advances in Neural Information Processing Systems (NeurIPS)},
  year={2021}
}

@inproceedings{gao2023pal,
  title={{PAL}: Program-Aided Language Models},
  author={Gao, Luyu and Madaan, Aman and Zhou, Shuyan and Alon, Uri and Liu, Pengfei and Yang, Yiming and Callan, Jamie and Neubig, Graham},
  booktitle={International Conference on Machine Learning (ICML)},
  year={2023}
}

@article{wan2024fusechat,
  title={{FuseChat}: Knowledge Fusion of Chat Models},
  author={Wan, Fanqi and Zhong, Longguang and Yang, Ziyi and Chen, Ruijun and Quan, Xiaojun},
  journal={arXiv preprint arXiv:2408.07990},
  year={2024}
}

@article{shen2024collm,
  title={{Co-LLM}: Learning to Decode Collaboratively with Multiple Language Models},
  author={Shen, Shannon Zejiang and Lang, Hunter and Wang, Bailin and Kim, Yoon and Sontag, David},
  journal={arXiv preprint arXiv:2403.03870},
  year={2024}
}

@article{zou2023representation,
  title={Representation Engineering: A Top-Down Approach to {AI} Transparency},
  author={Zou, Andy and Phan, Long and Chen, Sarah and Campbell, James and Guo, Phillip and Ren, Richard and Pan, Alexander and Yin, Xuwang and Mazeika, Mantas and Dombrowski, Ann-Kathrin and Goel, Shashwat and Li, Nathaniel and Byun, Michael J. and Wang, Zifan and Mallen, Alex and Basart, Steven and Koyejo, Sanmi and Song, Dawn and Fredrikson, Matt and Kolter, J. Zico and Hendrycks, Dan},
  journal={arXiv preprint arXiv:2310.01405},
  year={2023}
}

@inproceedings{todd2024function,
  title={Function Vectors in Large Language Models},
  author={Todd, Eric and Li, Millicent L. and Sharma, Arnab Sen and Mueller, Aaron and Wallace, Byron C. and Bau, David},
  booktitle={International Conference on Learning Representations (ICLR)},
  year={2024}
}

@inproceedings{pan2023logiclm,
  title={{Logic-LM}: Empowering Large Language Models with Symbolic Solvers for Faithful Logical Reasoning},
  author={Pan, Liangming and Albalak, Alon and Wang, Xinyi and Wang, William Yang},
  booktitle={Findings of the Association for Computational Linguistics: EMNLP},
  year={2023}
}

@inproceedings{du2024multiagent,
  title={Improving Factuality and Reasoning in Language Models through Multiagent Debate},
  author={Du, Yilun and Li, Shuang and Torralba, Antonio and Tenenbaum, Joshua B. and Mordatch, Igor},
  booktitle={International Conference on Machine Learning (ICML)},
  year={2024}
}

@inproceedings{wang2023selfconsistency,
  title={Self-Consistency Improves Chain of Thought Reasoning in Language Models},
  author={Wang, Xuezhi and Wei, Jason and Schuurmans, Dale and Le, Quoc and Chi, Ed and Narang, Sharan and Chowdhery, Aakanksha and Zhou, Denny},
  booktitle={International Conference on Learning Representations (ICLR)},
  year={2023}
}

@inproceedings{wei2022cot,
  title={Chain-of-Thought Prompting Elicits Reasoning in Large Language Models},
  author={Wei, Jason and Wang, Xuezhi and Schuurmans, Dale and Bosma, Maarten and Ichter, Brian and Xia, Fei and Chi, Ed and Le, Quoc and Zhou, Denny},
  booktitle={Advances in Neural Information Processing Systems (NeurIPS)},
  year={2022}
}

@article{bansal2024llmaugmented,
  title={{LLM} Augmented {LLMs}: Expanding Capabilities through Composition},
  author={Bansal, Rachit and Samanta, Bidisha and Dalmia, Siddharth and Gupta, Nitish and Vashishth, Shikhar and Ganapathy, Sriram and Bapna, Abhishek and Jain, Prateek and Talukdar, Partha},
  journal={arXiv preprint arXiv:2401.02412},
  year={2024}
}

@article{yang2024qwen25,
  title={Qwen2.5 Technical Report},
  author={Yang, An and Yang, Baosong and Zhang, Beichen and Hui, Binyuan and Zheng, Bo and others},
  journal={arXiv preprint arXiv:2412.15115},
  year={2024}
}

@article{yang2025qwen3,
  title={Qwen3 Technical Report},
  author={Yang, An and Li, Anfeng and Yang, Baosong and Zhang, Beichen and Hui, Binyuan and Zheng, Bo and Yu, Bowen and Gao, Chang and Huang, Chengen and Lv, Chenxu and others},
  journal={arXiv preprint arXiv:2505.09388},
  year={2025}
}

@inproceedings{demoura2008z3,
  title={{Z3}: An Efficient {SMT} Solver},
  author={de Moura, Leonardo and Bj{\o}rner, Nikolaj},
  booktitle={Tools and Algorithms for the Construction and Analysis of Systems (TACAS)},
  publisher={Springer},
  year={2008}
}

@misc{numina2024numinamath,
  title={{NuminaMath-TIR}},
  author={{AI-MO/NuminaMath-TIR contributors}},
  howpublished={HuggingFace Datasets: \texttt{AI-MO/NuminaMath-TIR}},
  year={2024}
}

@article{shao2024deepseekmath,
  title={{DeepSeekMath}: Pushing the Limits of Mathematical Reasoning in Open Language Models},
  author={Shao, Zhihong and Wang, Peiyi and Zhu, Qihao and Xu, Runxin and Song, Junxiao and Bi, Xiao and Zhang, Haowei and Zhang, Mingchuan and Li, Y.K. and Wu, Y. and Guo, Daya},
  journal={arXiv preprint arXiv:2402.03300},
  year={2024}
}

@inproceedings{ramesh2025activations,
  title={Communicating Activations Between Language Model Agents},
  author={Ramesh, Vignav and Li, Kenneth},
  booktitle={International Conference on Machine Learning (ICML)},
  year={2025}
}

@inproceedings{hao2024coconut,
  title={Training Large Language Models to Reason in a Continuous Latent Space},
  author={Hao, Shibo and Sukhbaatar, Sainbayar and Su, DiJia and Li, Xian and Hu, Zhiting and Weston, Jason and Tian, Yuandong},
  booktitle={Conference on Language Modeling (COLM)},
  year={2025}
}

@inproceedings{fu2026c2c,
  title={Cache-to-Cache: Direct Semantic Communication Between Large Language Models},
  author={Fu, Tianyu and Min, Zihan and Zhang, Hanling and Yan, Jichao and Dai, Guohao and Ouyang, Wanli and Wang, Yu},
  booktitle={International Conference on Learning Representations (ICLR)},
  year={2026}
}

@inproceedings{zheng2025thoughtcomm,
  title={Thought Communication in Multiagent Collaboration},
  author={Zheng, Yujia and Zhao, Zhuokai and Li, Zijian and Xie, Yaqi and Gao, Mingze and Zhang, Lizhu and Zhang, Kun},
  booktitle={Advances in Neural Information Processing Systems (NeurIPS)},
  year={2025}
}

@inproceedings{du2025interlat,
  title={Enabling Agents to Communicate Entirely in Latent Space},
  author={Du, Zhuoyun and Wang, Runze and Bai, Huiyu and Cao, Zouying and Zhu, Xiaoyong and Cheng, Yu and Zheng, Bo and Chen, Wei and Ying, Haochao},
  booktitle={Annual Meeting of the Association for Computational Linguistics (ACL)},
  year={2026}
}

@article{zou2025latentmas,
  title={Latent Collaboration in Multi-Agent Systems},
  author={Zou, Jiaru and Yang, Xiyuan and Qiu, Ruizhong and Li, Gaotang and Tieu, Katherine and Lu, Pan and Shen, Ke and Tong, Hanghang and Choi, Yejin and He, Jingrui and Zou, James and Wang, Mengdi and Yang, Ling},
  journal={arXiv preprint arXiv:2511.20639},
  year={2025}
}

@inproceedings{ye2023satlm,
  title={{SatLM}: Satisfiability-Aided Language Models Using Declarative Prompting},
  author={Ye, Xi and Chen, Qiaochu and Dillig, Isil and Durrett, Greg},
  booktitle={Advances in Neural Information Processing Systems (NeurIPS)},
  year={2023}
}

@inproceedings{olausson2023linc,
  title={{LINC}: A Neurosymbolic Approach for Logical Reasoning by Combining Language Models with First-Order Logic Provers},
  author={Olausson, Theo X. and Gu, Alex and Lipkin, Benjamin and Zhang, Cedegao E. and Solar-Lezama, Armando and Tenenbaum, Joshua B. and Levy, Roger},
  booktitle={Conference on Empirical Methods in Natural Language Processing (EMNLP)},
  year={2023}
}

@inproceedings{hao2023toolkengpt,
  title={{ToolkenGPT}: Augmenting Frozen Language Models with Massive Tools via Tool Embeddings},
  author={Hao, Shibo and Liu, Tianyang and Wang, Zhen and Hu, Zhiting},
  booktitle={Advances in Neural Information Processing Systems (NeurIPS)},
  year={2023}
}

@article{huh2024platonic,
  title={The Platonic Representation Hypothesis},
  author={Huh, Minyoung and Cheung, Brian and Wang, Tongzhou and Isola, Phillip},
  journal={arXiv preprint arXiv:2405.07987},
  year={2024}
}

@misc{anthropic2024claude35sonnet,
  title={Claude 3.5 Sonnet Model Card Addendum},
  author={Anthropic},
  year={2024},
  howpublished={\url{https://www-cdn.anthropic.com/fed9cc193a14b84131812372d8d5857f8f304c52/Model_Card_Claude_3_Addendum.pdf}}
}

@article{yang2026recursive,
  title={Recursive Multi-Agent Systems},
  author={Yang, Xiyuan and Zou, Jiaru and Pan, Rui and Qiu, Ruizhong and Lu, Pan and Diao, Shizhe and Jiang, Jindong and Tong, Hanghang and Zhang, Tong and Buehler, Markus J. and He, Jingrui and Zou, James},
  journal={arXiv preprint arXiv:2604.25917},
  year={2026}
}

@inproceedings{houlsby2019parameter,
  title={Parameter-Efficient Transfer Learning for {NLP}},
  author={Houlsby, Neil and Giurgiu, Andrei and Jastrzebski, Stanislaw and Morrone, Bruna and de Laroussilhe, Quentin and Gesmundo, Andrea and Attariyan, Mona and Gelly, Sylvain},
  booktitle={International Conference on Machine Learning},
  year={2019}
}

@article{wu2025dense,
  title={Dense Communication between Language Models},
  author={Wu, Shiguang and Wang, Yaqing and Yao, Quanming},
  journal={arXiv preprint arXiv:2505.12741},
  year={2025}
}

@article{dong2025parallel,
  title={Generalized Parallel Scaling with Interdependent Generations},
  author={Dong, Harry and Brandfonbrener, David and Helenowski, Eryk and He, Yun and Kumar, Mrinal and Fang, Han and Chi, Yuejie and Sankararaman, Karthik Abinav},
  journal={arXiv preprint arXiv:2510.01143},
  year={2025}
}

@inproceedings{dhanraj2025neurosymbolic,
  title={Improving Rule-based Reasoning in {LLMs} using Neurosymbolic Representations},
  author={Dhanraj, Varun and Eliasmith, Chris},
  booktitle={Conference on Empirical Methods in Natural Language Processing (EMNLP)},
  year={2025}
}

@article{turner2023actadd,
  title={Steering Language Models With Activation Engineering},
  author={Turner, Alexander Matt and Thiergart, Lisa and Leech, Gavin and Udell, David and Vazquez, Juan J. and Mini, Ulisse and MacDiarmid, Monte},
  journal={arXiv preprint arXiv:2308.10248},
  year={2023}
}

@inproceedings{li2023iti,
  title={Inference-Time Intervention: Eliciting Truthful Answers from a Language Model},
  author={Li, Kenneth and Patel, Oam and Vi{\'e}gas, Fernanda and Pfister, Hanspeter and Wattenberg, Martin},
  booktitle={Advances in Neural Information Processing Systems (NeurIPS)},
  year={2023}
}

@inproceedings{panickssery2024caa,
  title={Steering {Llama 2} via Contrastive Activation Addition},
  author={Rimsky, Nina and Gabrieli, Nick and Schulz, Julian and Tong, Meg and Hubinger, Evan and Turner, Alexander},
  booktitle={Proceedings of the 62nd Annual Meeting of the Association for Computational Linguistics (ACL)},
  year={2024}
}

@inproceedings{cunningham2024sparse,
  title={Sparse Autoencoders Find Highly Interpretable Features in Language Models},
  author={Cunningham, Hoagy and Ewart, Aidan and Riggs, Logan and Huben, Robert and Sharkey, Lee},
  booktitle={International Conference on Learning Representations (ICLR)},
  year={2024}
}

@misc{templeton2024scaling,
  title={Scaling Monosemanticity: Extracting Interpretable Features from {Claude 3 Sonnet}},
  author={Templeton, Adly and Conerly, Tom and Marcus, Jonathan and others},
  year={2024},
  howpublished={Transformer Circuits Thread, Anthropic},
  note={\url{https://transformer-circuits.pub/2024/scaling-monosemanticity/}}
}

@article{gao2024scalingsae,
  title={Scaling and Evaluating Sparse Autoencoders},
  author={Gao, Leo and la Tour, Tom Dupr{\'e} and Tillman, Henk and Goh, Gabriel and Troll, Rajan and Radford, Alec and Sutskever, Ilya and Leike, Jan and Wu, Jeffrey},
  journal={arXiv preprint arXiv:2406.04093},
  year={2024}
}

@inproceedings{marks2025featurecircuits,
  title={Sparse Feature Circuits: Discovering and Editing Interpretable Causal Graphs in Language Models},
  author={Marks, Samuel and Rager, Can and Michaud, Eric J. and Belinkov, Yonatan and Bau, David and Mueller, Aaron},
  booktitle={International Conference on Learning Representations (ICLR)},
  year={2025}
}

@inproceedings{fedus2022switch,
  title={Switch Transformers: Scaling to Trillion Parameter Models with Simple and Efficient Sparsity},
  author={Fedus, William and Zoph, Barret and Shazeer, Noam},
  booktitle={Journal of Machine Learning Research (JMLR)},
  year={2022}
}

@article{jiang2024mixtral,
  title={Mixtral of Experts},
  author={Jiang, Albert Q. and Sablayrolles, Alexandre and Roux, Antoine and Mensch, Arthur and Savary, Blanche and Bamford, Chris and Chaplot, Devendra Singh and de las Casas, Diego and Hanna, Emma Bou and Bressand, Florian and Lengyel, Gianna and Bour, Guillaume and Lample, Guillaume and Lavaud, L{\'e}lio Renard and Saulnier, Lucile and Lachaux, Marie-Anne and Stock, Pierre and Subramanian, Sandeep and Yang, Sophia and Antoniak, Szymon and Le Scao, Teven and Gervet, Th{\'e}ophile and Lavril, Thibaut and Wang, Thomas and Lacroix, Timoth{\'e}e and El Sayed, William},
  journal={arXiv preprint arXiv:2401.04088},
  year={2024}
}

@inproceedings{ilharco2023taskarithmetic,
  title={Editing Models with Task Arithmetic},
  author={Ilharco, Gabriel and Ribeiro, Marco Tulio and Wortsman, Mitchell and Gururangan, Suchin and Schmidt, Ludwig and Hajishirzi, Hannaneh and Farhadi, Ali},
  booktitle={International Conference on Learning Representations (ICLR)},
  year={2023}
}

@inproceedings{yadav2023ties,
  title={{TIES-Merging}: Resolving Interference When Merging Models},
  author={Yadav, Prateek and Tam, Derek and Choshen, Leshem and Raffel, Colin and Bansal, Mohit},
  booktitle={Advances in Neural Information Processing Systems (NeurIPS)},
  year={2023}
}

@inproceedings{yu2024dare,
  title={Language Models are Super {M}ario: Absorbing Abilities from Homologous Models as a Free Lunch},
  author={Yu, Le and Yu, Bowen and Yu, Haiyang and Huang, Fei and Li, Yongbin},
  booktitle={International Conference on Machine Learning (ICML)},
  year={2024}
}

@inproceedings{wortsman2022modelsoups,
  title={Model Soups: Averaging Weights of Multiple Fine-tuned Models Improves Accuracy Without Increasing Inference Time},
  author={Wortsman, Mitchell and Ilharco, Gabriel and Gadre, Samir Yitzhak and Roelofs, Rebecca and Gontijo-Lopes, Raphael and Morcos, Ari S. and Namkoong, Hongseok and Farhadi, Ali and Carmon, Yair and Kornblith, Simon and Schmidt, Ludwig},
  booktitle={International Conference on Machine Learning (ICML)},
  year={2022}
}

\appendix

\section{Extended related work}
\label{app:ext_related}

\textbf{Tool-augmented LLMs.}
Toolformer~\citep{schick2023toolformer}, ToRA~\citep{gou2024tora}, and PAL~\citep{gao2023pal} teach a single model to produce tool-calling syntax, modifying its weights. ToolkenGPT~\citep{hao2023toolkengpt} represents tools as learned token embeddings, enabling tool invocation through next-token prediction with frozen weights. The Bicameral Model takes a different approach: rather than embedding tool knowledge in one model, it routes tool access through a separate auxiliary model communicating via hidden states, keeping the primary model entirely unaware of tool syntax.

\textbf{Neurosymbolic reasoning.}
Logic-LM~\citep{pan2023logiclm}, SatLM~\citep{ye2023satlm}, and LINC~\citep{olausson2023linc} interface language models with formal solvers through text-level declarative specifications. LINC demonstrates that a 15B model with a first-order logic prover can surpass GPT-4 with chain-of-thought~\citep{wei2022cot}, the same pattern we observe with our 0.6B model and Z3. Closer to our framing, \citet{dhanraj2025neurosymbolic} embed an LM's activations into a structured neurosymbolic vector space so rule-based steps happen in latent symbolic form before being decoded back to tokens. Our approach differs from text-level solver interfaces in routing solver interaction through a hidden-state channel rather than text, and differs from in-LM neurosymbolic subspaces in pushing the symbolic component into a separate model (the auxiliary emits Z3 constraints).

\textbf{Hidden-state-level communication and coordination.}
Two recent lines of work share activations at the hidden-state level. \citet{wu2025dense} strip the token-level input/output layers from multiple LMs and connect them through trainable dense-communication modules, producing an end-to-end-differentiable graph of LMs that matches monolithic LM performance at a fraction of the training cost; the Bicameral Model preserves each model's token-level IO so the primary's output remains language, and couples per-token through a learned gate rather than Wu et al.'s continuous differentiable stream. \citet{dong2025parallel} share intermediate activations across parallel generations of the \emph{same} model via a learned bridge, turning independent samples into interdependent decoding paths (relatedly, Coconut~\citep{hao2024coconut} reasons in continuous latent space via hidden-state self-recurrence within one model); our coupling shares activations across models with complementary capabilities, while theirs shares activations across siblings for diversity.

\textbf{Activation steering and representation engineering.}
A parallel line of work edits LM behavior by adding fixed directions to intermediate hidden states. Representation engineering~\citep{zou2023representation} extracts concept directions that causally steer model outputs; function vectors~\citep{todd2024function} isolate task-specific directions that transfer across prompts; and contrastive activation addition~\citep{panickssery2024caa}, inference-time intervention~\citep{li2023iti}, and activation addition~\citep{turner2023actadd} intervene at inference with a static offset. The Bicameral Model's suppression gate can be viewed as a \emph{learned, token-level, bidirectional} relative of these static-direction steerings: the gate decides per-token whether and how much sender-derived signal to admit, rather than injecting a fixed vector uniformly across a sequence.

\textbf{Sparse autoencoders and feature circuits.}
A recent line of interpretability work trains sparse autoencoders on LM activations to decompose hidden states into monosemantic features~\citep{cunningham2024sparse,templeton2024scaling,gao2024scalingsae} and compose those features into causal circuits~\citep{marks2025featurecircuits}. We do not train SAEs, but this work is evidence that LM hidden states carry structured, manipulable features, which is the load-bearing assumption behind treating coupled hidden states as a usable communication medium.

\textbf{Mixture-of-experts.}
MoE architectures route each token to a small subset of specialized feed-forward experts within a single transformer~\citep{fedus2022switch,jiang2024mixtral}. The Bicameral Model shares the intuition that different parts of a computation are better handled by different specialized components, but differs in three ways: the specialized components are two full, separately pretrained models rather than expert FFNs inside one; coordination is bidirectional and continuous, not a per-token routing decision; and the auxiliary model runs its own decoding loop (including tool calls) rather than a feed-forward pass on the primary's token.

\textbf{Model merging and parameter composition.}
An alternative way to combine multiple pretrained models is to merge them at the weight level: task arithmetic~\citep{ilharco2023taskarithmetic}, TIES-Merging~\citep{yadav2023ties}, DARE~\citep{yu2024dare}, and model soups~\citep{wortsman2022modelsoups} average or combine parameters of models that share an architecture. These methods produce a single merged model and require architectural compatibility. Our coupling keeps both models' parameters intact, trains only a small interface between their activation spaces, and works with models of different sizes or architectures provided a translation network is learned.

\section{Experimental details}
\label{app:exp_details}

\paragraph{Arithmetic (Section~\ref{sec:arithmetic}).}
The primary and auxiliary models are both Qwen2.5-0.5B-Instruct~\citep{yang2024qwen25} (frozen, identical weights). The auxiliary model has access to a calculator tool. The interface uses \textsc{PullStandard} with a 3-hidden-layer translation MLP of hidden dimension 2048. Training follows a two-stage curriculum: 20 epochs of general arithmetic followed by 28 epochs of mixed arithmetic and GSM8K-type train data (Appendix~\ref{app:general_arithmetic}). Evaluation covers three benchmarks: general arithmetic (single binary expressions over $[0, 10^8]$ with log-uniform operand sampling; Appendix~\ref{app:general_arithmetic}), GSM8K~\citep{cobbe2021gsm8k} (grade-school word problems, 1{,}319 test examples), and GSM8K-IRL (professionally-themed transformations of GSM8K with realistic decimal quantities; Appendix~\ref{app:gsm8k_irl}).

\paragraph{Logic puzzles (Section~\ref{sec:zebra}).}
The primary and auxiliary models are both Qwen3-0.6B. The auxiliary model uses ZebraDSL (Appendix~\ref{app:zebradsl}), a DSL we designed for this purpose, backed by the Z3 constraint solver~\citep{demoura2008z3}. The interface uses \textsc{PullStandard} with coupling layers chosen proportionally to the best ranges found for the arithmetic problems. Training uses 10{,}000 procedurally generated logic puzzles per epoch, regenerated each epoch, with Z3-verified solutions. Puzzles span 2 to 6 entities and 2 to 6 attributes, covering 30 distinct clue types including direct constraints, indirect comparisons, and other patterns (Table~\ref{tab:zebradsl_clues}). We evaluate on ZebraLogic~\citep{lin2025zebralogic}, a benchmark of 1{,}000 logic grid puzzles ranging from 2$\times$2 to 6$\times$6 in size, and on GeneralZebra, a dynamically generated test suite of 1{,}000 puzzles matching the training distribution (Appendix~\ref{app:generalzebra}).
\section{Extended results}
\label{app:extended}

\subsection{ZebraLogic accuracy by puzzle size}
\label{app:zebralogic_by_size}

Table~\ref{tab:zebralogic_by_size} gives the full breakdown of ZebraLogic accuracy for the bicameral maximal configuration (Qwen3-0.6B + Z3, 40 epochs) across all 25 puzzle sizes from 2$\times$2 to 6$\times$6.

\begin{table}[h]
\centering
\caption{ZebraLogic accuracy by puzzle size ($N \times M$) for Bicameral (0.6B + Z3). Mean $\pm$ std over 5 seeds, each evaluated on $n=40$ puzzles per size.}
\label{tab:zebralogic_by_size}
\small
\begin{tabular}{cc|rr|l}
\toprule
$N$ & $M$ & Accuracy (\%) & Std (\%) & Group \\
\midrule
2 & 2 & 87.0 & 9.3 & Small \\
2 & 3 & 94.5 & 4.0 & Small \\
2 & 4 & 92.0 & 6.6 & Small \\
2 & 5 & 84.0 & 12.4 & Small \\
2 & 6 & 73.0 & 13.2 & Small \\
\midrule
3 & 2 & 72.5 & 20.1 & Small \\
3 & 3 & 93.5 & 3.4 & Small \\
3 & 4 & 74.0 & 6.4 & Medium \\
3 & 5 & 74.0 & 10.6 & Medium \\
3 & 6 & 56.0 & 15.5 & Medium \\
\midrule
4 & 2 & 65.5 & 18.5 & Small \\
4 & 3 & 81.5 & 3.4 & Medium \\
4 & 4 & 68.0 & 9.1 & Medium \\
4 & 5 & 55.0 & 5.7 & Large \\
4 & 6 & 43.5 & 13.2 & Large \\
\midrule
5 & 2 & 72.5 & 11.4 & Medium \\
5 & 3 & 66.5 & 8.5 & Large \\
5 & 4 & 51.0 & 3.4 & Large \\
5 & 5 & 41.0 & 12.0 & X-Large \\
5 & 6 & 36.0 & 6.4 & X-Large \\
\midrule
6 & 2 & 67.0 & 10.3 & Medium \\
6 & 3 & 62.5 & 7.1 & Large \\
6 & 4 & 49.0 & 9.3 & X-Large \\
6 & 5 & 35.5 & 7.3 & X-Large \\
6 & 6 & 22.0 & 3.7 & X-Large \\
\bottomrule
\end{tabular}
\vspace{0.5em}
\begin{tabular}{l|rr}
\toprule
Group & Accuracy (\%) & Std (\%) \\
\midrule
Small & 82.8 & 8.3 \\
Medium & 70.4 & 7.0 \\
Large & 55.7 & 5.3 \\
X-Large & 36.7 & 5.8 \\
\midrule
Overall & 64.7 & 6.0 \\
\bottomrule
\end{tabular}
\end{table}

\subsection{ZebraLogic training data evolution}
\label{app:version_progression}

Table~\ref{tab:version_progression} shows how ZebraLogic accuracy improved across training data versions. The architecture, interface, and model were held constant; only the training data generation was changed. The v2$\to$v4 improvement came from aligning the clue presentation format: ZebraLogic uses an ordinal framing (``There are $N$ houses, numbered 1 to $N$\ldots'') while GeneralZebra originally presented entities by name, creating a distribution mismatch the model could not bridge. The v4$\to$v5 improvement came from identifying 11 clue types present in ZebraLogic that were supported by ZebraDSL but had been overlooked in the GeneralZebra synthetic generator. These included double-indirect patterns where both sides reference entities by attribute value (e.g., ``the person who drinks coffee lives next to the person who owns a cat'') and mixed patterns combining one named entity with one indirect reference. These 11 types cover approximately 35\% of ZebraLogic clues that were nearly absent from earlier training data (Table~\ref{tab:zebradsl_clues}).

\begin{table}[h]
\centering
\caption{ZebraLogic accuracy across training data versions. All use Qwen3-0.6B with \textsc{PullStandard} interface (21M params), bic-maximal configuration (40 epochs). Only the training data generation was changed between versions. v2/v4 are single runs; v5 reports mean $\pm$ std over 5 seeds. GeneralZebra is regenerated for each version using the same generator as training, so its distribution evolves across rows (19 clue types in v2, 30 in v5); ZebraLogic is fixed across all versions.}
\label{tab:version_progression}
\small
\begin{tabular}{lccl}
\toprule
Version & ZebraLogic & GeneralZebra & Key change \\
\midrule
v2 & 4.1 & 96.1 & Baseline (19 clue types) \\
v4 & 20.5 & 91.3 & Format alignment (10+ fixes) \\
v5 & \textbf{64.7 $\pm$ 6.7} & \textbf{93.2 $\pm$ 0.9} & +11 double/mixed-indirect clue types \\
\bottomrule
\end{tabular}
\end{table}

\subsection{ZebraLogic failure mode analysis}
\label{app:failure_modes}

Analyzing the 248 failures from a single bic-maximal run (75.2\% accuracy, within the 5-seed distribution reported in Section~\ref{sec:zebra}), 65\% are DSL value errors (the auxiliary model uses entity or attribute names not matching the declared domain), 24\% are truncation (generation cut off before completing all constraints, concentrated on 5--6 entity puzzles), 9\% are solver-unsat (syntactically valid but logically contradictory constraints), and the remaining 2\% are other errors. The dominant error, hallucinated value names, reflects a grounding problem rather than a reasoning failure: the auxiliary model correctly identifies the constraint structure but fails to map natural-language values to their declared DSL identifiers. This suggests a clear path for improvement through training data normalization.

\subsection{ZebraDSL specification}
\label{app:zebradsl}

ZebraDSL is a declarative constraint language designed for logic grid puzzles. Commands are semicolon-delimited and processed through a five-stage pipeline (lexer $\to$ parser $\to$ semantic analyzer $\to$ Z3 translator $\to$ query executor). The solver maintains state across commands, allowing incremental constraint formulation. Each command receives an indexed acknowledgment (\texttt{=> [N];}), and queries return solver results. Below we describe the language primitives (Table~\ref{tab:zebradsl_syntax}) and the 30 clue types used in training data generation (Table~\ref{tab:zebradsl_clues}).

\paragraph{Declarations and queries.}
Entities are declared with \texttt{@entities:a,b,c;} and attribute domains with \texttt{@domain:color:red,blue,green;} (discrete) or \texttt{@domain:age:int[1,100];} (arithmetic). By default, all attributes are unique across entities; \texttt{@nonunique:hobby;} relaxes this. Queries take four forms: \texttt{?entity.attr;} (specific value), \texttt{?entity;} (all attributes), \texttt{?;} (satisfiability check), and \texttt{?json;} (full solution as JSON). Commands can be retracted with \texttt{!N;} (cancel command $N$) or \texttt{clear;} (reset all state).

\paragraph{Constraint syntax.}
Table~\ref{tab:zebradsl_syntax} lists the constraint primitives. Operands are either \emph{direct} (entity name, e.g., \texttt{alice.house}) or \emph{indirect} (attribute-value reference, e.g., \texttt{@pet:dog.house}, meaning ``the house number of whoever owns the dog''). All operators accept both operand forms, yielding three referencing modes: direct--direct, indirect--direct (or vice versa), and indirect--indirect.

\begin{table}[h]
\centering
\caption{ZebraDSL constraint syntax. Operands $L$ and $R$ are either direct (\texttt{entity.attr}) or indirect (\texttt{@attr:val.attr}) references. All constraints are semicolon-terminated.}
\label{tab:zebradsl_syntax}
\small
\begin{tabular}{lll}
\toprule
Operator & Syntax & Semantics \\
\midrule
Equality & \texttt{L=V;} or \texttt{L=R;} & Value assignment or same-value constraint \\
Inequality & \texttt{L!=V;} or \texttt{L!=R;} & Value exclusion or different-value constraint \\
Less than & \texttt{L<R;} & Strict ordering (ordinal attributes) \\
Greater than & \texttt{L>R;} & Strict ordering (ordinal attributes) \\
Disjunction & \texttt{L=V$_1$|V$_2$|$\cdots$;} & Value is one of the listed options \\
Adjacency & \texttt{L\textasciitilde$n$=R;} & $|L - R| = n$ (bidirectional distance) \\
Left-of & \texttt{L+1=R;} & $R = L + 1$ (directional, e.g., immediately left) \\
Right-of & \texttt{L-1=R;} & $R = L - 1$ (directional, e.g., immediately right) \\
Entity ref. & \texttt{@attr:val=entity;} & Bind indirect reference to named entity \\
\bottomrule
\end{tabular}
\end{table}

\paragraph{Clue types for training data.}
Training puzzles are generated procedurally with Z3-verified solutions. Each clue simultaneously produces a natural-language sentence (for the primary model's input) and one or more ZebraDSL commands (for the auxiliary model's target). Table~\ref{tab:zebradsl_clues} lists all 30 clue types organized by referencing mode. Clue types are sampled according to learned weights; \emph{attribute match} and \emph{attribute mismatch} are disabled by default as they are trivially satisfied (or impossible) under uniqueness constraints.

\begin{table}[h]
\centering
\caption{All 30 ZebraDSL clue types used in training data generation. \textbf{Direct} clues reference entities by name (\texttt{e.attr}); \textbf{indirect} clues reference entities by attribute value (\texttt{@a:v.attr}); \textbf{double-indirect} clues use indirect references on both sides; \textbf{mixed} clues combine one direct and one indirect operand. Notation: $e$ = entity name; $a$ = attribute; $v$ = value; $t$ = ordinal attribute.}
\label{tab:zebradsl_clues}
\footnotesize
\begin{tabular}{@{}rlll@{}}
\toprule
\# & Clue type & DSL pattern & NL example \\
\midrule
\multicolumn{4}{@{}l}{\textit{Direct (19 types)}} \\
1 & Equality & \texttt{e.a=v;} & ``Alice lives in house 2'' \\
2 & Inequality & \texttt{e.a!=v;} & ``Bob does not own a dog'' \\
3 & Less than & \texttt{e$_1$.a<e$_2$.a;} & ``Alice's house is left of Bob's'' \\
4 & Greater than & \texttt{e$_1$.a>e$_2$.a;} & ``Bob's house is right of Alice's'' \\
5 & Adjacency & \texttt{e$_1$.a\textasciitilde1=e$_2$.a;} & ``Alice and Bob are neighbors'' \\
6 & Between & \texttt{e$_1$.a<e$_2$.a; e$_2$.a<e$_3$.a;} & ``Bob is between Alice and Carol'' \\
7 & Attr.\ match$^\dagger$ & \texttt{e$_1$.a=e$_2$.a;} & ``Alice and Bob have the same hobby'' \\
8 & Attr.\ mismatch$^\dagger$ & \texttt{e$_1$.a!=e$_2$.a;} & ``Alice and Bob have different hobbies'' \\
9 & OR constraint & \texttt{e.a=v$_1$|v$_2$;} & ``Alice's pet is a dog or a cat'' \\
10 & Entity ref. & \texttt{@a:v=e;} & ``The Norwegian is Alice'' \\
11 & Indirect eq. & \texttt{@a$_1$:v$_1$.a$_2$=v$_2$;} & ``The dog owner lives in house 3'' \\
12 & Indirect $<$ & \texttt{@a:v.t<e.t;} & ``The dog owner is left of Bob'' \\
13 & Indirect $>$ & \texttt{@a:v.t>e.t;} & ``The dog owner is right of Bob'' \\
14 & Indirect $\neq$ & \texttt{@a$_1$:v$_1$.a$_2$!=v$_2$;} & ``The dog owner is not Norwegian'' \\
15 & Dir.\ left & \texttt{e$_1$.t+1=e$_2$.t;} & ``Alice is directly left of Bob'' \\
16 & Dir.\ right & \texttt{e$_1$.t-1=e$_2$.t;} & ``Alice is directly right of Bob'' \\
17 & Indir.\ adj. & \texttt{@a$_1$:v$_1$.t\textasciitilde1=@a$_2$:v$_2$.t;} & ``The Norwegian and dog owner are neighbors'' \\
18 & Distance 2 & \texttt{e$_1$.t\textasciitilde2=e$_2$.t;} & ``One house between Alice and Bob'' \\
19 & Distance 3 & \texttt{e$_1$.t\textasciitilde3=e$_2$.t;} & ``Two houses between Alice and Bob'' \\
\midrule
\multicolumn{4}{@{}l}{\textit{Double-indirect (6 types) --- both operands are \texttt{@a:v} references}} \\
20 & Indir.\ $<$ & \texttt{@a$_1$:v$_1$.t<@a$_2$:v$_2$.t;} & ``The Norwegian is left of the dog owner'' \\
21 & Indir.\ $>$ & \texttt{@a$_1$:v$_1$.t>@a$_2$:v$_2$.t;} & ``The Norwegian is right of the dog owner'' \\
22 & Indir.\ left & \texttt{@a$_1$:v$_1$.t+1=@a$_2$:v$_2$.t;} & ``The Norwegian is directly left of the dog owner'' \\
23 & Indir.\ right & \texttt{@a$_1$:v$_1$.t-1=@a$_2$:v$_2$.t;} & ``The Norwegian is directly right of the dog owner'' \\
24 & Indir.\ dist.\ 2 & \texttt{@a$_1$:v$_1$.t\textasciitilde2=@a$_2$:v$_2$.t;} & ``One house between the Norwegian and dog owner'' \\
25 & Indir.\ dist.\ 3 & \texttt{@a$_1$:v$_1$.t\textasciitilde3=@a$_2$:v$_2$.t;} & ``Two houses between the Norwegian and dog owner'' \\
\midrule
\multicolumn{4}{@{}l}{\textit{Mixed (5 types) --- one direct, one indirect operand}} \\
26 & Mixed left & \texttt{e.t+1=@a:v.t;} & ``Alice is directly left of the dog owner'' \\
27 & Mixed right & \texttt{e.t-1=@a:v.t;} & ``Alice is directly right of the dog owner'' \\
28 & Mixed adj. & \texttt{e.t\textasciitilde1=@a:v.t;} & ``Alice and the dog owner are neighbors'' \\
29 & Mixed dist.\ 2 & \texttt{e.t\textasciitilde2=@a:v.t;} & ``One house between Alice and the dog owner'' \\
30 & Mixed dist.\ 3 & \texttt{e.t\textasciitilde3=@a:v.t;} & ``Two houses between Alice and the dog owner'' \\
\bottomrule
\end{tabular}

\vspace{0.3em}
{\footnotesize $^\dagger$Disabled by default (trivially true or impossible under uniqueness constraints).}
\end{table}

\subsection{GeneralZebra evaluation benchmark}
\label{app:generalzebra}

GeneralZebra is a dynamically generated test suite of 1{,}000 logic grid puzzles drawn from the same distribution as the training data. It measures in-distribution generalization and complements ZebraLogic, which tests out-of-distribution transfer.

\textbf{Puzzle generation procedure.}
Each puzzle is constructed in three stages: (1)~a random solution is created by sampling entities from one of 22 categories, sampling attributes and domain values, then randomly assigning values to entities; (2)~clues are added one at a time (sampled from the 30 clue types weighted by configuration) until the constraint set yields a unique solution as verified by Z3; (3)~a greedy minimization pass removes clues iteratively, restarting from the first clue after each successful removal, until no single clue can be dropped without compromising uniqueness. The result is a puzzle that is locally minimal in clue count. Table~\ref{tab:generalzebra_config} lists the generation parameters.

\begin{table}[h]
\centering
\caption{GeneralZebra generation parameters. Unless noted, values are shared with the training data generator.}
\label{tab:generalzebra_config}
\small
\begin{tabular}{ll}
\toprule
Parameter & Value \\
\midrule
Number of puzzles & 1{,}000 \\
Entities per puzzle & 2--6 (uniform) \\
Attributes per puzzle & 2--6 (uniform) \\
Ordinal attribute ratio & 0.5 \\
Entity categories & All 22 (incl.\ zebralogic-style) \\
Clue types & All 30 (default weights) \\
Solution uniqueness & Z3-verified \\
Clue minimization & Greedy redundant-clue removal \\
Query type & Mixed: 40\% single-attribute, 30\% multiple-attribute, 30\% json\_table \\
Validation & Cell-wise match (json\_table) or exact value match (single/multiple) \\
\bottomrule
\end{tabular}
\end{table}

\textbf{Relation to training data.}
GeneralZebra puzzles are produced by the same generator code as training puzzles. Isolation relies on fresh random generation at evaluation time (1{,}000 puzzles) versus fresh generation at each training epoch (10{,}000 puzzles per epoch, 40 epochs). Given the enormous combinatorial puzzle space (22 entity categories, variable dimensions, random solutions and constraints), structural overlap is vanishingly unlikely, though train and test sets are not explicitly seed-partitioned. JSON answer validation follows the same pipeline as ZebraLogic: case-insensitive, separator-normalized comparison of every entity--attribute--value cell, with a puzzle scored correct only when all cells match.

\textbf{Interpretation.}
The low variance across seeds on GeneralZebra ($93.2 \pm 0.9$\%) compared to ZebraLogic ($64.7 \pm 6.7$\%) reflects the in-distribution nature of the benchmark. GeneralZebra is best interpreted as a measure of whether the interface has learned to coordinate constraint formulation reliably, while ZebraLogic measures transfer to a fixed benchmark with different formatting conventions and entity categories.

\subsection{Adapter-equivalent ablation design}
\label{app:adapter_equiv}

The adapter-equivalent ablation isolates the contribution of the auxiliary model by retaining the interface parameters but bypassing $M_a$ entirely. This section formalizes the signal flow.

\paragraph{Signal path.}
In the full bicameral system, reverse coupling (Eq.~\ref{eq:coupling_reverse}) updates the primary's hidden state at $\ell_w^{a \to p}$ using the auxiliary's hidden state at $\ell_r^{a \to p}$. In adapter-equivalent mode, $M_a$ is never executed. Instead, the primary model's read-layer activations are passed through both translation networks in series (first $f^{p \to a}$, then $f^{a \to p}$), creating a round-trip perturbation that replaces the auxiliary-derived signal in Eq.~\ref{eq:coupling_reverse}:
\begin{equation}
\label{eq:adapter_equiv}
\mathbf{h}_p^{(\ell_w^{a \to p})}(t) \;\gets\; \big(1 - \sigma^{a \to p}(t)\big)\, \mathbf{h}_p^{(\ell_w^{a \to p})}(t) \;+\; \sigma^{a \to p}(t) \, f^{a \to p}\!\Big(f^{p \to a}\!\big(\mathbf{h}_p^{(\ell_r^{p \to a})}(t)\big)\Big).
\end{equation}
The suppression gate $\sigma^{a \to p}(t) = \mathrm{Sigmoid}\!\left(g^{a \to p}(\mathbf{h}_p^{(\ell_w^{a \to p})}(t))\right)$ is unchanged; it still reads the primary model's write-layer state. Architecturally, the round-trip $f^{a \to p} \circ f^{p \to a}$ composes the two translation MLPs into a bottleneck autoencoder ($d \to 2048 \to 2048 \to d \to 2048 \to 2048 \to d$) on the primary model's own representations. The forward coupling direction ($M_p \to M_a$) is unused since there is no auxiliary model to receive it.

\paragraph{Training protocol.}
The adapter-equivalent is trained \emph{from scratch} with the auxiliary model bypassed throughout, not evaluated post-hoc on a bicameral checkpoint. All hyperparameters, curriculum stages, data generation, and epoch counts are identical to the corresponding bicameral configuration. Only the primary model's cross-entropy loss provides gradient signal; this loss flows backward through the frozen primary layers, through $f^{a \to p}$, and through $f^{p \to a}$, so both translation networks receive gradient updates despite the auxiliary model being absent. The total trainable parameter count is identical to the full bicameral system.

\paragraph{Why this controls for parameter count.}
The adapter-equivalent tests whether the interface parameters alone, without an auxiliary model generating tool outputs, can learn a useful transformation. The interface has the same capacity (21M parameters for Qwen3-0.6B) and training signal as the bicameral system. The observed result (7.5\% on ZebraLogic versus 64.7\% bicameral, and 48.0\% versus 96.5\% on arithmetic) confirms that the gains are driven by the auxiliary model's reasoning, not by the interface acting as a parameter-efficient adapter.

\subsection{Dual-target loss and masking}
\label{app:masking}

Section~\ref{sec:dual_target_sft} states that the training loss is $\mathcal{L} = \mathcal{L}_p + \mathcal{L}_a$ with prompt and tool-output tokens masked. Here we detail the full masking system and loss normalization.

\paragraph{Loss computation.}
Each loss term is computed as a \emph{masked mean}: per-token cross-entropy is multiplied element-wise by a mask tensor $\mathbf{m} \in [0,1]^T$, then summed and normalized by the mask's active mass:
\[
\mathcal{L}_p = \frac{\sum_{t=1}^{T} m_t^{(p)} \cdot \ell_t^{(p)}}{\sum_{t=1}^{T} m_t^{(p)}}, \qquad
\mathcal{L}_a = \frac{\sum_{t=1}^{T} m_t^{(a)} \cdot \ell_t^{(a)}}{\sum_{t=1}^{T} m_t^{(a)}}
\]
where $\ell_t$ is the per-token cross-entropy at position $t$. The two terms are summed with \emph{equal weight} (no tunable coefficient). Because each is independently normalized by its own active token count, the two loss scales are comparable even when the number of active positions differs substantially between the primary and auxiliary streams.

\paragraph{Masking categories.}
Masks are continuous-valued tensors constructed by the causality constraint solver. They are initialized to 1 (all tokens active) and then modified by six categories, applied in order:

\begin{enumerate}[leftmargin=*,itemsep=2pt]
\item \textbf{Primary prompt} ($m^{(p)} \gets 0$). All input prompt tokens are masked so that only the primary model's response tokens contribute to $\mathcal{L}_p$.

\item \textbf{Aux prompt} ($m^{(a)} \gets 0$). The auxiliary model's system prompt is masked, including one additional token beyond the prompt boundary. This extra token is masked because coupling activates at the first generated auxiliary token, not the last prompt token.

\item \textbf{Forced tool output} ($m^{(a)} \gets 0$). Tokens injected by tool execution (e.g., calculator results \texttt{=478272;} or Z3 solver responses \texttt{=> [3];}) are masked individually. The auxiliary model did not generate these tokens, so training on them would be inappropriate and would waste interface capacity on learning to predict externally-determined outputs.

\item \textbf{Aux dropout ranges} ($m^{(a)} \gets 0$). When auxiliary content blocks are stochastically dropped during data construction (controlled by an auxiliary dropout probability, default 0), the token range where the dropped content would have appeared is masked. This prevents the model from learning to predict wait tokens where content was expected.

\item \textbf{Mirrored prompt tokens} ($m^{(a)} \gets 0$). When the primary model's prompt tokens are copied into the auxiliary stream (mirror configuration), these forced tokens are masked from $\mathcal{L}_a$. Note: mirror configurations were explored but are not used in any experiments reported in this paper. All reported results use a no-mirror setup where the auxiliary receives only its tool-instruction system prompt and must infer task content entirely through the hidden-state channel. Mirror configurations may be useful in practice to reduce the burden on the interface for initial value transfer.

\item \textbf{Wait-token downweighting} ($m^{(a)} \gets w$, where $w \in [0,1]$). Applied last. Non-content auxiliary tokens that survive the above categories (predominantly wait tokens between tool calls) have their mask value reduced from 1 to a configurable weight $w$. Content tokens (tool calls, DSL commands) retain mask value 1. This focuses $\mathcal{L}_a$ on meaningful generation rather than predicting long stretches of wait tokens. The default is $w = 1$ (no downweighting).
\end{enumerate}

Categories 1--5 set mask values to 0 (binary exclusion). Category~6 is the only source of fractional mask values. The multiplication in the loss formula means that a token with $m_t = 0.01$ contributes 1\% as much gradient as a token with $m_t = 1.0$.

\subsection{Auxiliary dropout}
\label{app:aux_dropout}

During training data construction for the arithmetic domain, each auxiliary content block (e.g., a \texttt{calc(...)} call) is independently dropped with probability $p$. When a block is dropped, its token positions are filled with wait tokens and masked from $\mathcal{L}_a$ (category 4 above). The primary model's loss and token sequence are unchanged.

The motivation is primary-model robustness. Without dropout, if the auxiliary model fails to produce a calculator call at inference time (due to out-of-distribution inputs or gate suppression), the primary model has never encountered the absence of a reverse-coupling signal at positions where it was trained to expect one. In practice, this causes the primary model to emit space tokens indefinitely, waiting for a response that never arrives. By occasionally omitting auxiliary content during training, the primary model learns to recognize when no response is forthcoming and proceed with its best estimate.

All arithmetic experiments reported in this paper use $p = 0.004$ (approximately 1 in 250 content blocks dropped per sample). This value was selected from a sweep over $[0, 0.112]$; higher values degraded auxiliary model performance while lower values provided insufficient robustness. The ZebraLogic experiments do not use auxiliary dropout ($p = 0$).

\subsection{Training data generation pipeline}
\label{app:data_pipeline}

The causality-tagged training data for word problems (GSM8K and GSM8K-IRL) is produced by a three-stage pipeline that transforms the target model's own outputs into corrected, annotated examples.

\paragraph{Stage 1: Harvesting.}
The target primary model (Qwen2.5-0.5B-Instruct) is run on GSM8K train (7{,}473 problems) and GSM8K-IRL train (7{,}473 problems) with temperature 0.1 and max generation length 200 tokens. On GSM8K train, the model answers 71.8\% correctly; on GSM8K-IRL train, only 13.6\% are correct. All outputs (correct and incorrect) are retained for annotation.

\paragraph{Stage 2: Minimal correction and tool annotation.}
Claude~3.5 Sonnet~\citep{anthropic2024claude35sonnet} processes each harvested response in a single pass. The prompt instructs Claude to:
\begin{itemize}[nosep,leftmargin=*]
    \item Identify errors in the model's reasoning and apply minimal corrections, preserving the original style, tone, and phrasing as much as possible.
    \item Generate \texttt{calc(expression)} entries (restricted to $+, -, \times, \div$ and parentheses) for each arithmetic step.
    \item Insert causality tags at their appropriate positions: \texttt{@@CALC1@@} at the earliest point where the calculation can be inferred from context, and \texttt{@@RESPONSE1@@} immediately before the numerical result first appears.
\end{itemize}
For correct responses, Claude still inserts calculator expressions and tags without modifying the text. A verification variant re-evaluates the generated \texttt{calc(...)} expressions with Python and gives Claude the exact results, allowing it to re-correct if its own arithmetic was wrong.

\paragraph{Stage 3: Validation.}
The annotated outputs are validated programmatically: \texttt{calc(...)} expressions must parse correctly and contain only digits and basic operators; tags must appear in correct order; and the final extracted answer must match the ground-truth (with tolerance for formatting differences). Records failing validation are retried (up to 3 attempts) or discarded.

\paragraph{Success rates.}
The pipeline achieves 99.1\% success on GSM8K and 98.6\% on GSM8K-IRL (after retries). The high success rate on GSM8K reflects that most responses are already correct and require only tag insertion. GSM8K-IRL's lower initial correctness rate (13.6\%) means more corrections are needed, but Claude's single-pass correction is reliable for arithmetic word problems.

\paragraph{Design rationale.}
Using the target model's own outputs (rather than a reference model's solutions) as the starting point is deliberate: the corrected text remains in-distribution for the primary model's generation style, reducing the distribution shift between training data and inference. The minimal-correction strategy further preserves this property: Claude changes only what is mathematically wrong, leaving phrasing, step ordering, and vocabulary intact. This means the primary model trains on text it might plausibly have generated itself, with calculator results available through the coupling. This principle is important because the interface must learn to steer the primary model's own generation patterns rather than imitate a different model's style. A more principled variant of this pipeline, in which Claude makes atomic single-error fixes and the target model regenerates all subsequent tokens (ensuring $>$95\% of training tokens are the target model's own output), is under development for future work.

\subsection{Causality-constrained data alignment}
\label{app:causality}

Section~\ref{sec:causality} describes the causality constraint system at a high level. Here we detail the data structures, constraint solving algorithm, and tag anchoring patterns.

\paragraph{Problem statement.}
Both models must produce token sequences of equal length for lockstep alignment. The auxiliary model's content (tool calls and their results) must be placed so that: (a) the auxiliary model does not act on information the primary model has not yet produced at that token position (causal ordering), and (b) tool results are available before the primary model needs them (timing). Between active tool calls, the auxiliary stream is filled with wait tokens.

\paragraph{Tag anchoring.}
Each training example embeds \emph{causality tags} (markers of the form \texttt{@@TAG\_NAME@@}) directly into the primary model's response text. Each tag marks a logical boundary, such as the end of a question or the start of an answer. An auxiliary content block specifies two tags: an \texttt{after\_tag} (the block must not start until the primary model has passed this position) and a \texttt{before\_tag} (the block must complete before the primary model reaches this position). The tag processor resolves each tag to a token index by: (1) recording the tag's character position in the tagged text, (2) stripping all tags to produce clean text, (3) tokenizing the prefix up to the adjusted character position and counting tokens. The clean text (without tags) is what both models actually see during training and inference.

\paragraph{Constraint solver.}
Given a list of auxiliary content blocks with their tag-derived placement windows, the solver walks the blocks sequentially, maintaining a cursor position. For each block, it:
\begin{enumerate}[leftmargin=*,itemsep=1pt]
\item Tokenizes the auxiliary content and any forced tool output to compute token lengths.
\item Computes the valid placement window: $[\,\max(\textit{after\_pos}, \textit{cursor}),\; \textit{before\_pos} - \textit{content\_len}\,]$.
\item Selects a position within this window using the scheduling strategy (see below).
\item Checks for constraint violations: a \emph{before-violation} if the block's end exceeds \textit{before\_pos} (tool result arrives too late), or an \emph{after-violation} if the block's start precedes \textit{after\_pos} (causal ordering broken).
\item If violated, applies a fallback policy: \texttt{allow} (keep despite violation), \texttt{drop\_ar\_output} (skip this block), \texttt{drop\_sample} (discard the entire example), or \texttt{primary\_wait} (insert space tokens into the primary sequence to create room; see below).
\item Advances the cursor past the placed content and forced output.
\end{enumerate}

\paragraph{Scheduling strategies.}
Four strategies determine where within a valid window the auxiliary content is placed:

\begin{itemize}[leftmargin=*,itemsep=1pt]
\item \textbf{Eager}: earliest valid position, $\max(\textit{after\_pos}, \textit{cursor})$. The auxiliary model fires its tool call as soon as it causally can.
\item \textbf{Lazy}: latest valid position, $\textit{before\_pos} - \textit{content\_len}$, clamped to the window floor. The auxiliary model waits until the last possible moment.
\item \textbf{Random}: uniform sample within the valid window. This is used in all reported experiments.
\item \textbf{Balanced}: identical to random in the current implementation.
\end{itemize}

\noindent Random placement provides data augmentation: the same logical example produces different token-level alignments across epochs (combined with per-epoch regeneration of the underlying problems).

\paragraph{Primary-wait fallback.}
When a before-violation occurs (the auxiliary content does not fit before the \texttt{before\_tag} position), the \texttt{primary\_wait} policy \emph{modifies the primary token sequence} to create room. It calculates the number of extra tokens needed ($\textit{required\_end} - \textit{before\_pos}$) and inserts that many space tokens into the primary stream at the constraint boundary. All subsequent positions shift accordingly. The effect: the primary model emits extra space tokens to ``wait'' for the auxiliary model to finish its tool call. This is the only policy that alters the primary model's training data. In our experiments, before-violations use \texttt{primary\_wait}; after-violations (causal ordering) use \texttt{drop\_sample}.

\paragraph{Worked example: arithmetic.}
Consider the problem ``What is $564 \times 848$?'' with answer 478{,}272. Tags are placed in the primary response as follows:

\vspace{0.5em}
\noindent\begin{minipage}{\linewidth}
\small
\textbf{Primary (tagged):} \texttt{564 * 848 equals {\color{blue}@@ANSWER\_READY@@}478272.}\\[3pt]
\textbf{Input (tagged):} \texttt{What is 564 {\color{blue}@@QUESTION\_END@@}* 848?}\\[3pt]
\textbf{Aux content:} \texttt{calc(564*848)} with forced output \texttt{ =478272;}\\
\hspace*{2em}\texttt{after\_tag}: \textsc{question\_end} \quad \texttt{before\_tag}: \textsc{answer\_ready}
\end{minipage}
\vspace{0.5em}

\noindent The constraint solver resolves \textsc{question\_end} to the token index of ``564'' and \textsc{answer\_ready} to the token index of ``478272''. The auxiliary block \texttt{calc(564*848)} must be placed after the primary has produced the first operand (so the auxiliary model has ``seen'' it via hidden-state coupling) and before the primary needs the answer. Between these positions, the auxiliary stream contains wait tokens; at the selected position, the auxiliary stream contains \texttt{calc(564*848)=478272;} (content + forced output, the latter masked from loss).

\paragraph{Worked example: logic puzzle.}
For zebra puzzles, the think block uses a chain of tags that anchor each DSL command to its corresponding natural-language restatement:

\vspace{0.5em}
\noindent\begin{minipage}{\linewidth}
\small
\begin{tabular}{@{}p{0.48\linewidth}@{\hspace{0.5em}}p{0.48\linewidth}@{}}
\textbf{Primary think block (abridged)} & \textbf{Aux stream} \\[3pt]
\texttt{{\color{blue}@@ENT@@}Entities: Alice, Bob, Carol} & \texttt{@entities:alice,bob,carol; {\color{gray}=> [1];}} \\[1pt]
\texttt{{\color{blue}@@A1@@}- Pet: dog, cat, fish} & \texttt{@domain:pet:dog,cat,fish; {\color{gray}=> [2];}} \\[1pt]
\texttt{{\color{blue}@@A2@@}- House: 1, 2, 3} & \texttt{@domain:house:1,2,3; {\color{gray}=> [3];}} \\[1pt]
\texttt{{\color{blue}@@C1@@}Alice lives in house 2} & \texttt{alice.house=2; {\color{gray}=> [4];}} \\[1pt]
\texttt{{\color{blue}@@C2@@}Bob does not own a dog} & \texttt{bob.pet!=dog; {\color{gray}=> [5];}} \\[1pt]
\texttt{{\color{blue}@@SQ@@}...solver informs me...} & \texttt{?json; {\color{gray}=> \{...\};}} \\[1pt]
\texttt{{\color{blue}@@SR@@}\{solution JSON\}} & \\
\end{tabular}
\end{minipage}
\vspace{0.5em}

\noindent Each auxiliary command's placement window is bounded by adjacent tags: the entity declaration is placed between \textsc{ent} and \textsc{a1}, the first domain between \textsc{a1} and \textsc{a2}, the first constraint between \textsc{c1} and \textsc{c2}, and the \texttt{?json;} query between \textsc{sq} and \textsc{sr}. This chaining ensures monotonic ordering: every DSL command appears in the auxiliary stream at a position where the primary model has just restated the corresponding piece of the puzzle, and before it moves on to the next. Gray text shows forced tool output (Z3 acknowledgments and results), which is masked from the auxiliary loss.

\subsection{Reinforcement learning on the interface}
\label{app:rl}

After supervised fine-tuning, the interface parameters can be further optimized with reinforcement learning. We implement GRPO~\citep{shao2024deepseekmath} with the following design:

\begin{itemize}[leftmargin=*,itemsep=2pt]
\item \textbf{What is optimized.} Only the neural interface parameters $\theta_\phi$ (including the suppression gate). Both LLMs remain frozen and in eval mode.
\item \textbf{Trajectory generation.} For each prompt, $K$ trajectories (default 96) are generated through the full coupled pipeline with temperature sampling, producing diverse primary--auxiliary token sequences.
\item \textbf{Reward.} Outcome-based: binary 1/0 for answer correctness. An optional tool-use reward component scores whether the auxiliary model invoked tools appropriately, though empirically this has minimal effect beyond the outcome signal.
\item \textbf{Advantage computation.} Per-prompt group normalization: rewards are mean-centered and std-normalized within each prompt's trajectory group. A subset (default 16) is selected for training, balancing high-reward and low-reward trajectories.
\item \textbf{Policy gradient.} PPO-style clipped objective ($\epsilon = 0.2$) over the token distributions of both models, with DAPO-style loss normalization. Gradients flow through both models' forward passes but only update the interface.
\end{itemize}

\textbf{Preliminary results.}
On the arithmetic domain (starting from an SFT-trained \textsc{PullStandard} checkpoint), RL training yields consistent improvements on held-out word problems: GSM8K-IRL accuracy rises from 16.1\% (SFT) to 20.8\% (+4.7pp), with no regression on calculator-based arithmetic (96--97\%) or GSM8K (39--43\%). Results are reproducible: four identical RL runs show standard deviations below 1\% on all benchmarks. We swept learning rate (1e-6 to 3e-3), sampling temperature, reward weighting, and trajectory selection strategy across 150+ runs; the optimal learning rate (3e-6) is an order of magnitude below SFT, and performance is robust to other hyperparameter choices.

\textbf{RL as a timing optimizer.}
SFT requires carefully engineered causality constraints (Appendix~\ref{app:causality}) to align the two token streams, and the quality of this alignment affects performance. RL sidesteps this: because it optimizes the interface end-to-end on outcome reward, it can learn when the suppression gate should fire and how strongly, effectively discovering coordination timing from the reward signal rather than from hand-crafted placement windows. We leave a full investigation of RL training, including application to the ZebraLogic domain and scaling to larger models, to future work.

\subsection{Python tool coupling examples}
\label{app:python_examples}

Using twin Qwen3-4B models with an identity interface (trained on NuminaMath-TIR~\citep{numina2024numinamath}), the auxiliary model receives no problem text, only hidden-state signals from the primary model at intermediate layers. Despite this, it generates problem-specific Python code.

\textbf{Training data construction.}
The 72k NuminaMath-TIR examples are split into two streams without correctness filtering. Each example's serial solution (interleaved reasoning text and \texttt{python}/\texttt{output} blocks) is parsed into segments. The primary stream receives all reasoning text wrapped in a think block, with causality tags inserted where code blocks originally appeared. The auxiliary stream receives the extracted code blocks, with execution outputs attached as forced tokens (masked from loss). Causality constraints anchor each code block between tags so it is placed after the primary has produced the relevant reasoning context and before the primary needs the result.

\textbf{Results.}
Aggregate accuracy (62.5\% on MATH) does not exceed the unaugmented Qwen3-4B with thinking enabled (81.6\% on MATH at 32k tokens). However, on 127 MATH problems (2.5\% of the benchmark), the bicameral system produces correct answers where thinking alone fails. Below are examples illustrating what the hidden-state channel transmits; full coupling traces are shown in Figures~\ref{fig:full_python_gcd}--\ref{fig:full_python_remainder}.

\textbf{Example 1: Modular arithmetic (wins vs.\ thinking baseline).}
Problem: ``Tim is doing a cakewalk with 11 steps. He takes his first step on step~1 and takes a total of 139 steps, walking in a circle. Which step does he end on?'' The unaugmented Qwen3-4B with 32k-token thinking gets this wrong.

\vspace{-0.5em}
\begin{verbatim}
  final_step = (1 + 139 - 1) % 11
  print(final_step)  # 7
\end{verbatim}
\vspace{-0.5em}

Three lines. The auxiliary recovered both key numbers (11, 139) and the correct offset formula from hidden states alone. Answer: correct.

\textbf{Example 2: Implicit algebraic simplification.}
Problem: ``Let $n = 2^4 \cdot 3^5 \cdot 4^6 \cdot 6^7$. How many natural-number factors does $n$ have?''

\vspace{-0.5em}
\begin{verbatim}
  exponent_2 = 23
  exponent_3 = 12
  num_factors = (exponent_2 + 1) * (exponent_3 + 1)
  print(num_factors)  # 312
\end{verbatim}
\vspace{-0.5em}

The auxiliary emitted the consolidated prime exponents (23, 12) rather than the raw values (4, 5, 6, 7), performing the simplification $4^6 = 2^{12}$, $6^7 = 2^7 \cdot 3^7$ implicitly. Answer: correct.

\textbf{Example 3: Base conversion (wins vs.\ thinking baseline).}
Problem: ``Captain Rusczyk tracked down a pirate who had stolen $2345_6$ dollars. After winning a duel, he demands $41324_5$ dollars. How much has the pirate gone in debt? Express in base~10.'' The unaugmented model gets this wrong.

\vspace{-0.5em}
\begin{verbatim}
  base_6 = 2*(6**3) + 3*(6**2) + 4*(6**1) + 5*(6**0)
  base_5 = 4*(5**4) + 1*(5**3) + 3*(5**2) + 2*(5**1) + 4*(5**0)
  print(base_6 - base_5)  # -2145
\end{verbatim}
\vspace{-0.5em}

The auxiliary reconstructed both digit strings (\texttt{2345} in base~6, \texttt{41324} in base~5) and both bases from hidden states. Answer: correct (debt = 2145).

\textbf{Example 4: Recursive sequence with 7 recovered parameters.}
Problem: ``Given $x_1{=}211, x_2{=}375, x_3{=}420, x_4{=}523$ and $x_n = x_{n-1} - x_{n-2} + x_{n-3} - x_{n-4}$, find $x_{531} + x_{753} + x_{975}$.''

\vspace{-0.5em}
\begin{verbatim}
  x = [0] * 1000
  x[1]=211; x[2]=375; x[3]=420; x[4]=523
  for n in range(5, 976):
      x[n] = x[n-1] - x[n-2] + x[n-3] - x[n-4]
  print(x[531] + x[753] + x[975])  # 898
\end{verbatim}
\vspace{-0.5em}

All four initial values, the alternating-sign recurrence, and the three target indices were recovered from hidden states. Answer: correct.

\subsection{Layer-wiring constraints and inference}
\label{app:wiring_constraints}

\paragraph{Acyclicity constraint.}
The coupling configuration is defined by four layer indices: the primary read layer $\ell_r^p$, primary write layer $\ell_w^p$, auxiliary read layer $\ell_r^a$, and auxiliary write layer $\ell_w^a$. Since activations flow upward through layers (from layer 0 to layer $L$), cross-model coupling must not introduce a cycle in the activation dependency graph. The four coupling operations must admit a topological sort that respects layer ordering within each model.

The coupling order is determined by the primary model's layer relationship. When $\ell_r^p \leq \ell_w^p$ (\textbf{standard order}), the primary is read first (clean, before receiving any perturbation), its signal is written to the auxiliary at $\ell_w^a$, the auxiliary continues to $\ell_r^a$ and is read, and that signal is written back to the primary at $\ell_w^p$. When $\ell_w^p < \ell_r^p$ (\textbf{reversed order}), the auxiliary is read first (clean), written to the primary, the primary continues to its read layer, and that signal is written back to the auxiliary.

Within each model, read and write layers can independently take any relative ordering (read $<$ write, read $=$ write, or write $<$ read), giving 9 combinations. However, one combination is invalid: when the primary uses reversed order ($\ell_w^p < \ell_r^p$), the auxiliary must have $\ell_r^a < \ell_w^a$ (strict). Otherwise, the primary-to-auxiliary perturbation is injected at a layer that is never used for the auxiliary's final forward pass, causing it to be silently lost. All other 8 combinations are valid. When read $=$ write on a given model, the operation order follows the coupling direction: the model that goes first reads clean then receives its write; the model that goes second receives its write then is read (picking up the perturbed state).

The 890-configuration sweep samples from this space of valid configurations.

\paragraph{KV cache with perturbed activations.}
During inference, each model maintains a KV cache for efficient autoregressive decoding. When the interface injects a perturbation at a write layer, the perturbed hidden state continues through the remaining layers and is stored in the KV cache at each subsequent layer. This means future tokens attend to the perturbed representations when computing attention over past positions. The coupling effect thus persists across time steps through the cache, even though the interface only operates on the current token's hidden states. Position IDs are derived from the cache sequence length to ensure correct positional encoding during incremental decoding.

\subsection{Ablations and scaling}
\label{app:ablations}

All experiments use AdamW with learning rate $4 \times 10^{-4}$, gradient clipping at 1.0, and batch sizes of 2--8 (with gradient accumulation where noted). The learning rate was not tuned per experiment. Table~\ref{tab:interface_configs} lists the interface configurations evaluated.

\begin{table}[h]
\centering
\caption{Interface configurations. All use ReLU activations and identical twin models (same weights for primary and auxiliary). Adapters use $\alpha{=}128$ where present.}
\label{tab:interface_configs}
\scriptsize
\begin{tabular}{@{}llllrll@{}}
\toprule
Domain & Model ($d$) & Interface & Params & Translation & Suppression & Adapters \\
\midrule
Arithmetic & Qwen2.5-0.5B (896) & \textsc{PullStd} & 16M & 3L, $h{=}2048$ & 3L, $h{=}64$ & None \\
Arithmetic & Qwen2.5-0.5B (896) & \textsc{ScalarId} & 803K & Identity & 2L, $h{=}64$ & r64, L7/10/15 \\
Arithmetic & Qwen2.5-0.5B (896) & \textsc{ElemId} & 919K & Identity & 2L, $h{=}64$ & r64, L7/10/15 \\
ZebraLogic & Qwen3-0.6B (1536) & \textsc{PullStd} & 21M & 3L, $h{=}2048$ & 3L, $h{=}64$ & None \\
Python tool & Qwen3-4B (2560) & \textsc{ScalarId} & 2.3M & Identity & 2L, $h{=}64$ & r64, L11/15/23 \\
\bottomrule
\end{tabular}
\end{table}

\textbf{Coupling depth.}
We swept five same-depth coupling configurations (layers 6/18, 8/16, 10/15, 10/20, 12/14) across both identity architectures with and without adapters (40 configurations total). All mean accuracies in this section are averaged across general arithmetic, GSM8K, and GSM8K-IRL test sets. Mid-layer coupling at depths 10--15 and 10--20 is consistently optimal, achieving 54.9\% and 51.9\% mean accuracy respectively. Shallow coupling (6/18, 8/16) degrades all metrics, falling to 25--31\%. This pattern is consistent across interface types.

\textbf{Adapter placement.}
With identity interfaces, adapter placement relative to the coupling layers is the dominant factor. Across 32 configurations testing eight placement strategies at two coupling depths, adapters at the coupling layers produce the best combined results (46.7\% mean across all three benchmarks), while early-layer adapters (layers 3/5/7) are worse than no adapters at all (22.0\% versus 34.5\%). Adapter rank 32--64 is the sweet spot; the suppression network size has little effect.

\textbf{Curriculum effect.}
The arithmetic warm-up phase is essential for word-problem transfer on harder problems. Training 48 epochs on GSM8K + GSM8K-IRL alone (no arithmetic stage) achieves comparable GSM8K accuracy (41.7\% vs 39.9\% with curriculum) but nearly halves GSM8K-IRL performance (11.8\% vs 21.6\%). Without arithmetic practice, the auxiliary model still learns to trigger \texttt{calc()} expressions (78--100\% trigger rate, 99\% syntactically valid), but the primary model fails to reliably integrate the calculator results into its reasoning through the coupling. The pure arithmetic stage teaches the interface to faithfully transfer quantities between the two models' hidden states, while the word-problem examples teach the system \emph{when} to invoke the calculator. GSM8K-IRL's realistic decimal quantities amplify this effect: its sub-computations require precise multi-step decimal arithmetic where the primary model struggles to succeed without reliable calculator integration.

\textbf{Robustness to layer configuration.}
In a separate sweep of 890 layer wiring configurations using \textsc{PullStandard}, 95\% of configurations improved over the primary-only arithmetic baseline and 23.7\% improved on GSM8K-IRL, confirming that the coupling mechanism is robust to exact layer placement within the mid-to-late optimal region.

\textbf{Model scaling.}
Preliminary experiments confirm the architecture functions at larger scales, including asymmetric pairings (32B primary with 0.6B auxiliary using pipeline parallelism). However, configurations were not optimized at larger scales, and word-problem degradation persisted, likely due to our initial experiments reusing annotated training data initially harvested from the Qwen2.5-0.5B model. Characterizing scaling behavior and emergent properties of the bicameral configuration is a direction for future work.

\subsection{Layer-wiring sweep}
\label{app:layer_sweep}

\begin{figure}[h]
\centering
\includegraphics[width=\linewidth]{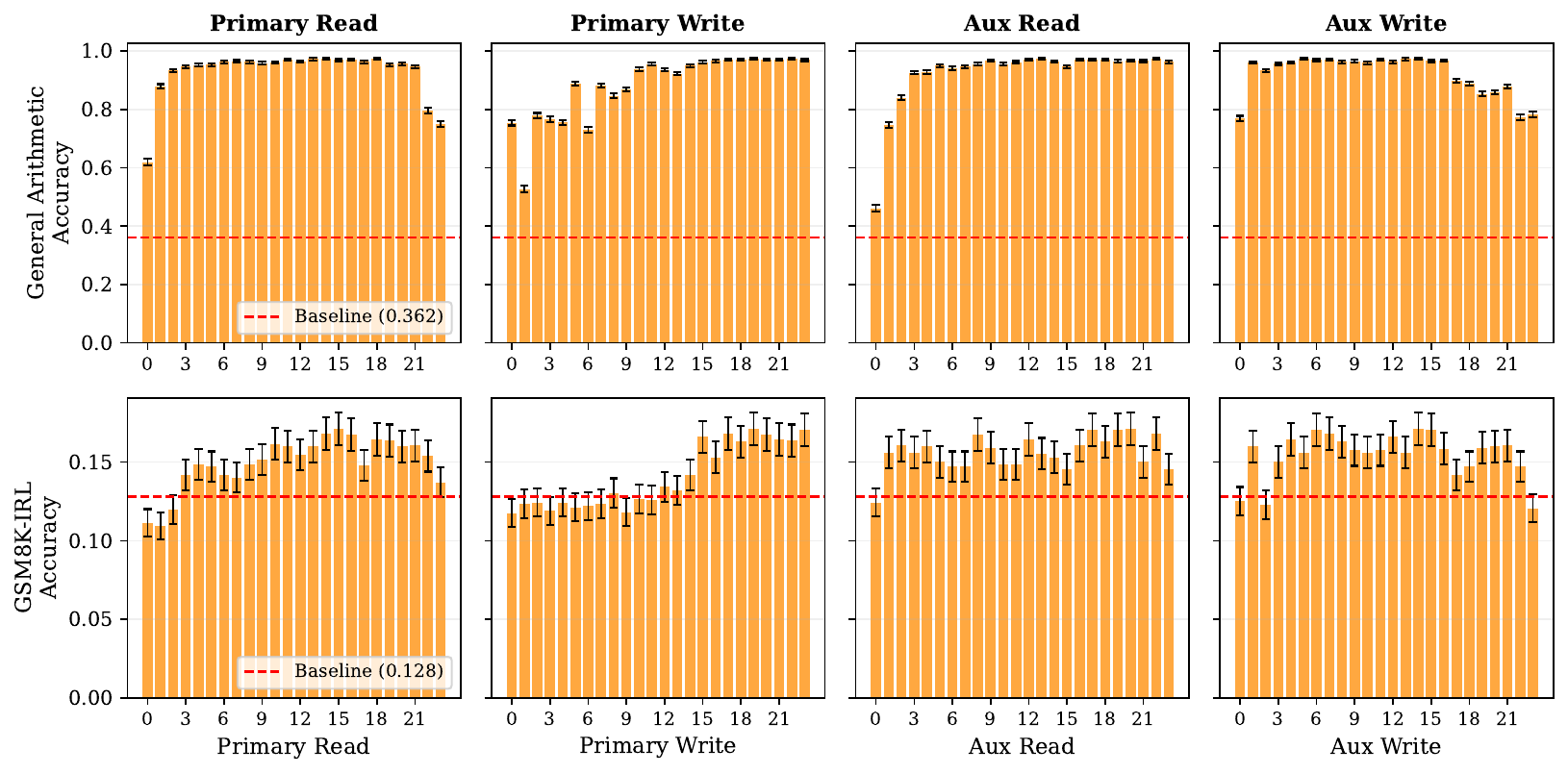}
\caption{Best performance for each layer value across 890 configurations, marginalized over other layer choices. \textbf{Top:} General arithmetic. \textbf{Bottom:} GSM8K-IRL. Red dashed line: primary-only baseline. Mid-to-late layers (10--20) consistently achieve the highest ceilings. Error bars: Bernoulli SEM.}
\label{fig:layer_sweep_max}
\end{figure}

The 890-configuration sweep varies which transformer layers are tapped for coupling in both models across all valid coupling-order regimes. Figure~\ref{fig:layer_sweep_max} shows the best accuracy achieved by any configuration using each layer value (marginalized over all other layer choices). Error bars show Bernoulli standard error ($\sqrt{p(1-p)/n}$) from the finite test set, reflecting measurement uncertainty.

\textbf{Statistical significance methodology.}
To determine whether a configuration significantly outperforms the primary-only baseline, we use a two-sample test based on the Bernoulli standard error. For a configuration with accuracy $p_c$ on $n_c$ samples and a baseline with accuracy $p_b$ (averaged over 8 runs, SEM $s_b$): the combined uncertainty is $s = \sqrt{s_b^2 + p_c(1-p_c)/n_c}$, and a configuration is classified as significantly better if $(p_c - p_b) > 2s$ (approximately 95\% confidence). Of the 890 configurations, 95.4\% improve over the primary-only baseline on general arithmetic ($n=2{,}000$), of which 94.7\% (843 configs) are statistically significant. On GSM8K-IRL ($n=1{,}319$), 23.7\% improve over the baseline, of which 6.6\% (59 configs) reach significance. On GSM8K ($n=1{,}319$), 0\% improve.

\begin{figure}[h]
\centering
\includegraphics[width=\linewidth]{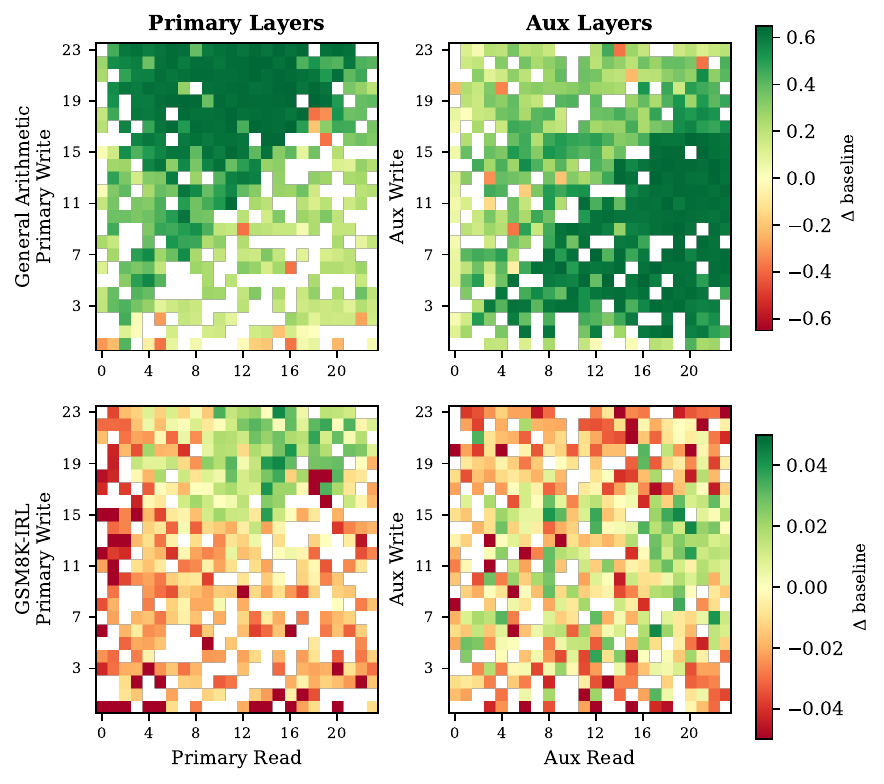}
\caption{Best improvement over primary-only baseline for each projected layer pair across 890 configurations. \textbf{Top:} General arithmetic. \textbf{Bottom:} GSM8K-IRL. Green regions contain configurations that substantially beat the baseline; the optimal region (mid-to-late layers for both read and write) is clearly visible. White cells have no data for that layer pair.}
\label{fig:layer_sweep_baseline}
\end{figure}

Figure~\ref{fig:layer_sweep_baseline} shows the best improvement over the primary-only baseline for each projected layer pair, revealing which layer regions contain at least one configuration that substantially outperforms the baseline.

\textbf{Interpreting the heatmap structure.}
The heatmaps in Figure~\ref{fig:layer_sweep_baseline} reveal clear triangular structure that aligns with the architecture's causal constraints. For primary model layers, the high-performance region forms an upper-left triangle corresponding to primary read $<$ primary write (standard ordering, where the primary model is read first). For auxiliary model layers, the high-performance region forms a lower-right triangle corresponding to auxiliary write $<$ auxiliary read. Together, these define the ``standard ordering with round-trip within one token'' regime: the primary is read, its signal is written to the auxiliary at an earlier layer, the auxiliary processes this signal through its intermediate layers, and its response is read at a later layer to be written back to the primary, all within a single generation step.

This is precisely the configuration depicted in Figure~\ref{fig:system} and the regime we initially hypothesized would be optimal before implementing the sweep. Configurations outside this triangle still work (the acyclicity constraint is satisfied), but the round-trip is broken: if the auxiliary is read before being written to (auxiliary read $<$ auxiliary write), both couplings still fire within the same token, but the reverse coupling cannot be informed by the latest forward coupling. The auxiliary's response at its read layer reflects only prior tokens' forward signals, not the current one. The sweep confirms that this broken round-trip is harmful but not fatal.

The heatmaps also reveal a clear asymmetry between standard and reversed coupling order. Reversed order (primary write $<$ primary read, where the auxiliary is read first) performs consistently worse, even within the round-trip regime. The likely explanation is that reversed order has the auxiliary model ``drive'' the interaction: its hidden states influence the primary model before the primary's own representations can inform the auxiliary about the current task context. Since the primary model sees the actual problem and the auxiliary does not, this ordering is counterproductive. In this task setup, the auxiliary likely benefits more from responding to the primary's signal than from leading the interaction.

Within the optimal triangular region, performance concentrates further in a specific subregion: mid-to-late primary read (layers 10--15) paired with a primary write a few layers beyond, and mid-to-late auxiliary write paired with a later auxiliary read. This pattern is consistent across both general arithmetic and GSM8K-IRL, suggesting that the optimal coupling taps representations after the models have built sufficient contextual understanding (past the early layers) but before the final layers commit to specific token predictions.

\subsection{General arithmetic benchmark}
\label{app:general_arithmetic}

The general arithmetic evaluation set is generated synthetically with the same procedure used for training data, ensuring no overlap between training and evaluation (different random seeds, regenerated each epoch during training). Each problem is a single binary expression $a \circ b$ where $\circ \in \{+, -, \times, \div\}$.

\textbf{Number sampling.} Operands are drawn from $[0, 10^8]$ using a mixed sampling strategy: with probability 0.05 the value is sampled uniformly, otherwise log-uniformly ($10^{U(0, 8)}$). Log-uniform sampling ensures coverage across magnitudes (from single-digit to hundred-million-scale) while the small uniform component prevents complete absence of very large values. Each operand is then rounded to a randomly selected precision level (a power of 10 from $10^{-2}$ to $10^3$, sampled uniformly). The precision controls decimal granularity, not magnitude: a large sample like 45{,}283{,}899.78 becomes 45{,}284{,}000 at precision $10^3$, 45{,}283{,}900 at precision $10^1$, or remains 45{,}283{,}899.78 at precision $10^{-2}$. The minimum representable value is clamped to the precision factor. Operands are negated with probability 0.01.

\textbf{Operator handling.} All four operators are equally likely. Division avoids zero denominators; subtraction swaps operands to produce positive results. Answers are computed by the calculator tool and verified to arbitrary precision ($\leq$5 decimal places in the formatted output).

\textbf{Template formatting.} Problems are presented in natural language using one of seven template categories (basic, conversational, \LaTeX, formal, casual, mathematical, verbose), with 90\% basic templates (e.g., ``Calculate 423 + 15{,}000'') and the remainder distributed across other styles. This variety teaches the primary model to recognize arithmetic requests in diverse phrasings.

\textbf{Evaluation protocol.} Each evaluation set consists of 10{,}000 problems generated with a fixed seed (42). The model receives the problem text and must produce the correct numerical answer. We report exact-match accuracy after normalizing formatting (commas, trailing zeros).

\textbf{Training curriculum.} The 48-epoch curriculum uses two stages: (1) epochs 0--20: 20{,}000 pure arithmetic problems regenerated each epoch; (2) epochs 20--48: a mixed dataset of ${\sim}$15{,}800 samples (46.7\% GSM8K-IRL, 46.9\% GSM8K, 6.3\% general arithmetic), with the arithmetic portion regenerated each epoch. The transition to mixed data teaches the interface to route word problems to the calculator while preserving the primary model's chain-of-thought reasoning.

\subsection{GSM8K-IRL benchmark}
\label{app:gsm8k_irl}

GSM8K-IRL (``In Real Life'') is a synthetically generated benchmark that transforms GSM8K problems into harder, professionally-themed variants with realistic non-round quantities. We generate it using Claude~3.5 Sonnet~\citep{anthropic2024claude35sonnet} via a structured transformation pipeline.

\textbf{Generation.} Each of the 7{,}473 GSM8K training problems (and separately, the 1{,}319 test problems) is sent to Claude~3.5 Sonnet with a transformation prompt instructing it to:
\begin{itemize}[nosep,leftmargin=*]
    \item Replace child/school contexts with adult professional scenarios (workplace, finance, engineering, science, legal, etc.)
    \item Replace simple round numbers with realistic decimal quantities (e.g., \$20 $\to$ \$19.89)
    \item Maintain the same number of calculation steps and the same arithmetic operators ($+, -, \times, \div$ only)
    \item Produce a step-by-step solution and a final extracted numerical answer
\end{itemize}

\textbf{Verification.} A separate verification pass ensures solution correctness. Claude extracts all arithmetic sub-expressions from the generated solution, Python \texttt{eval()} computes each expression exactly, and Claude reviews the full solution given the exact calculation results to determine whether the final answer is correct. Problems with incorrect solutions are flagged.

\textbf{Error rate.} The realistic decimal quantities introduce systematic rounding and truncation errors at a ${\sim}$2\% rate (155 suspect ground-truth labels out of 7{,}473 training items), approximately 14$\times$ higher than GSM8K's ${\sim}$0.15\% error rate. These errors arise because multi-step decimal arithmetic is genuinely harder, and the generation model occasionally rounds intermediate results. We do not filter these from the training set, as the 2\% noise rate has negligible impact on training signal.

\textbf{Dataset statistics.} The train split contains 7{,}473 problems (one per GSM8K train example) and the test split contains 1{,}319 problems (one per GSM8K test example). Problems span diverse professional domains including finance, engineering, chemistry, logistics, and project management.

\textbf{Evaluation.} We evaluate by extracting the numerical answer from model output (last \texttt{\textbackslash boxed\{\}} match, or last number in the final sentence) and comparing against the ground truth using \texttt{math.isclose} with relative tolerance $10^{-5}$ and absolute tolerance $10^{-8}$. This is effectively exact match for integer answers but accommodates minor floating-point discrepancies on decimal results. The test split is a transfer benchmark: training uses the IRL train split as part of the mixed-data curriculum (stage 2), while evaluation uses the held-out IRL test split derived from different seed problems.

\subsection{Compute resources}
\label{app:compute}

Table~\ref{tab:compute} summarizes GPU-hours across all experiments reported in this paper. The total is approximately 26{,}000 GPU-hours, predominantly on NVIDIA L40S (48GB) and H200 (80GB) GPUs.

\begin{table}[h]
\centering
\caption{Compute resources by experiment category.}
\label{tab:compute}
\small
\begin{tabular}{lrrrl}
\toprule
Category & Runs & GPUs/run & Hours/run & GPU-hours \\
\midrule
Arithmetic layer sweep (890 configs) & 890 & 1 L40S & $\sim$17 & 15{,}130 \\
Arithmetic identity/adapter sweeps & 230 & 1 L40S & 7--17 & 2{,}760 \\
Arithmetic multiseed (2$\times$5 seeds) & 10 & 1 L40S & $\sim$17 & 170 \\
ZebraLogic training (v2--v5, L40S) & 30 & 1 L40S & 4--12 & 180 \\
ZebraLogic v5 multiseed (5 seeds) & 5 & 1 H200 & $\sim$55 & 275 \\
ZebraLogic v5 extended (80--160 ep) & 5 & 1 H200 & $\sim$110 & 550 \\
ZebraLogic adapter-equivalent & 2 & 1 H200 & $\sim$55 & 110 \\
Python tool / TIR (Qwen3-4B) & 27 & 8 L40S/H200 & $\sim$11 & 2{,}376 \\
Communication onset (train + eval) & 106 & 1 L40S & 1--8 & 320 \\
RL experiments & 150 & 1 L40S & $\sim$6 & 900 \\
Multi-GPU scaling (8B, 32B) & 4 & 8 L40S & $\sim$80 & 2{,}560 \\
Evaluation (all benchmarks) & 200+ & 1--8 & 1--2 & 500 \\
\midrule
\textbf{Total} & & & & $\mathbf{\sim}$\textbf{26{,}000} \\
\bottomrule
\end{tabular}
\end{table}

Training times per run ranged from 4 hours (20-epoch ZebraLogic on H200) to 18 hours (48-epoch arithmetic curriculum on L40S). The 890-experiment layer sweep dominates total compute (58\%). All sub-1B experiments used single GPUs; the 4B Python-tool and 8B/32B scaling experiments used 8-GPU pipeline parallelism.

\subsection{Full coupling visualizations}
\label{app:full_visualizations}

The following figures show complete token-by-token coupling traces for successful generations across all three tool domains. Each visualization displays the full primary and auxiliary token streams with coupling strength at every position, revealing the temporal coordination patterns the interface learns from task loss alone. In all figures, \textbf{blue} indicates forward coupling ($M_p \to M_a$), \textbf{red} indicates reverse coupling ($M_a \to M_p$), and \textbf{green} highlights forced output tokens where tool results are injected into the auxiliary's hidden-state stream.

\subsubsection{Arithmetic with calculator}

\begin{figure*}[h]
\centering
\includegraphics[width=\linewidth]{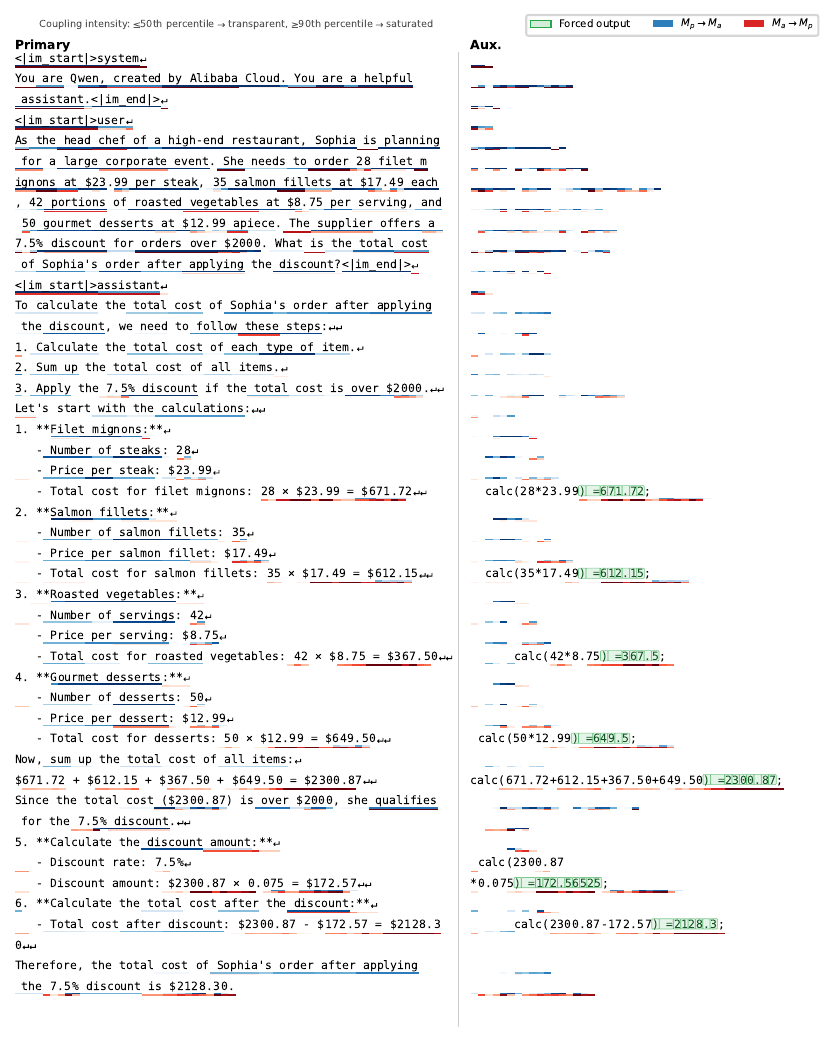}
\caption{%
\textbf{Full coupling trace: multi-step arithmetic word problem} (Qwen2.5-0.5B-Instruct, scalar identity interface, calculator tool).
The auxiliary (right column) waits with spaces until the primary emits relevant tokens, then issues sequential \texttt{calc(...)} calls.
Forward coupling (blue) is sustained while the primary emits dollar amounts and quantity words;
reverse coupling (red) is near-zero except at forced output positions (green, \texttt{=671.72}, \texttt{=612.15}, etc.) where calculator results flow back, or are recalled when the primary model references those values later (e.g., for the final answer).
The primary model incorporates each result into its chain-of-thought with single-token latency.
Figure~\ref{fig:coupling_activity} in the main text shows a zoomed excerpt of this trace (tokens 464--515).
}
\label{fig:full_arithmetic}
\end{figure*}

Figure~\ref{fig:full_arithmetic} shows the complete generation for a multi-step word problem requiring five arithmetic operations and a conditional discount.

\subsubsection{Logic puzzle with Z3 solver}

\begin{figure*}[h]
\centering
\includegraphics[width=\linewidth]{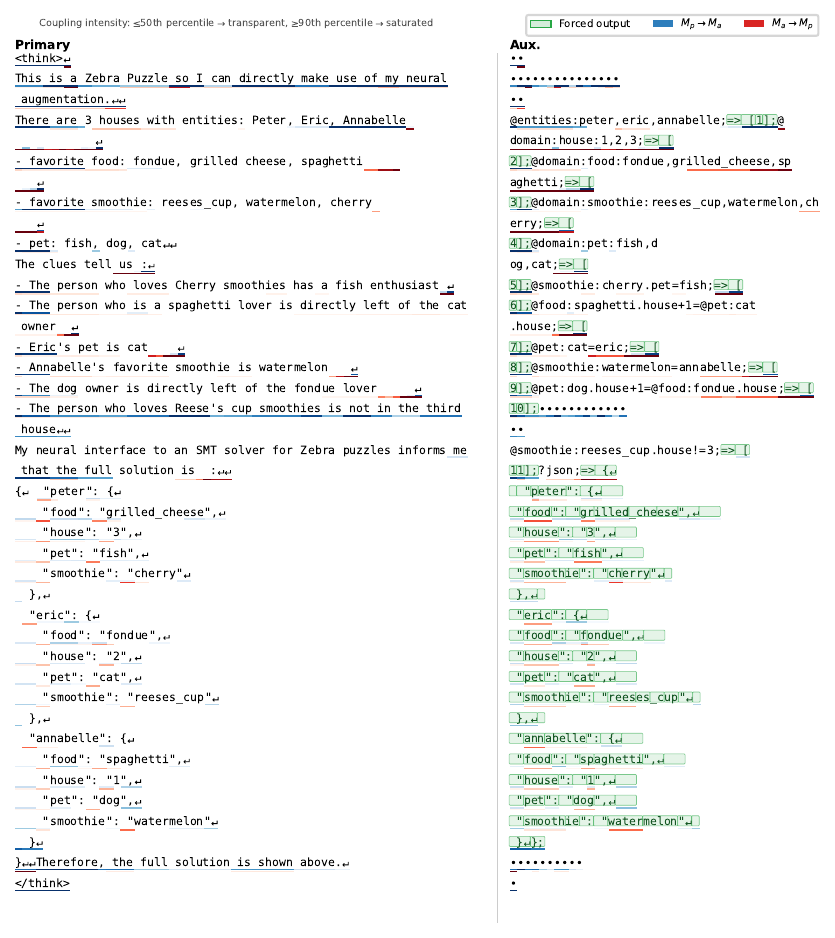}
\caption{%
\textbf{Full coupling trace: 3$\times$3 logic puzzle} (Qwen3-0.6B, standard interface, ZebraDSL/Z3 tool).
This puzzle is modified from a ZebraLogic benchmark instance: ``Arnold'' $\to$ ``Annabelle,'' ``pizza'' $\to$ ``fondue,'' and ``desert'' $\to$ ``Reese's cup.'' None of these replacement values appear anywhere in the training data across any entity category, testing out-of-distribution generalization.
The auxiliary (right column) emits wait tokens (\texttt{\textbullet}) during Phase~2 while it processes the problem through the forward channel, then produces a burst of ZebraDSL constraint declarations (\texttt{@entities}, \texttt{@domain}, \texttt{@smoothie:cherry.pet=fish}, etc.) during Phase~3.
Coupling strengths show that the auxiliary attends primarily to the first presentation of each clue during Phase~2; when the primary reiterates clues in its thought block, the auxiliary registers only the first few tokens before emitting its translation.
Forced output tokens (green) carry the solver's JSON solution back through the reverse channel.
The primary model's output (left column) transitions from internal reasoning to the correct final JSON assignment.
}
\label{fig:full_zebra}
\end{figure*}

Figure~\ref{fig:full_zebra} shows the complete generation for a 3$\times$3 ZebraLogic puzzle with 4 attributes and 6 clues.

\begin{figure*}[h]
\centering
\includegraphics[width=\linewidth]{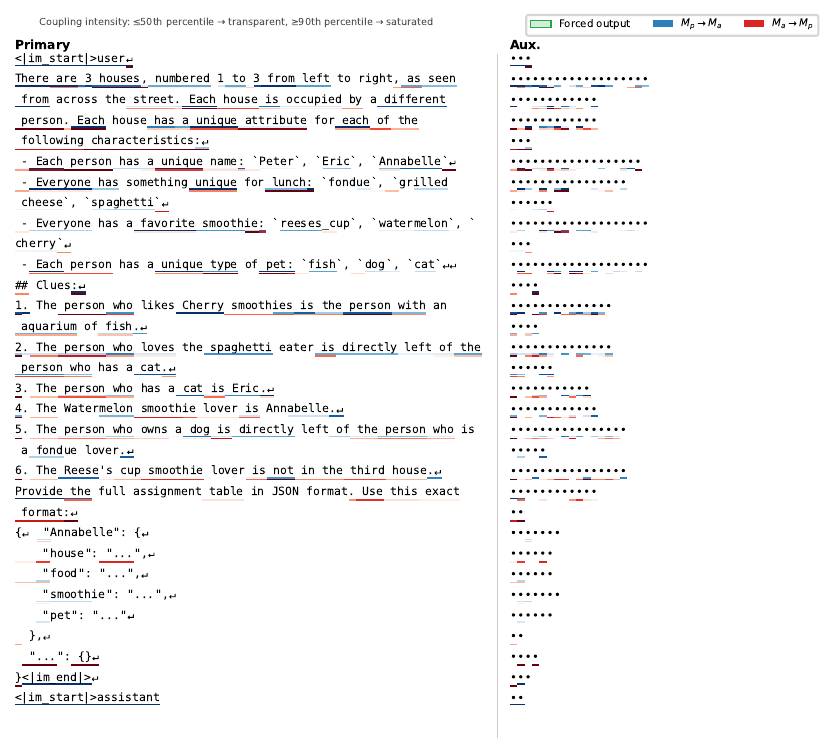}
\end{figure*}

\subsubsection{Python tool}

\begin{figure*}[h]
\centering
\includegraphics[width=\linewidth]{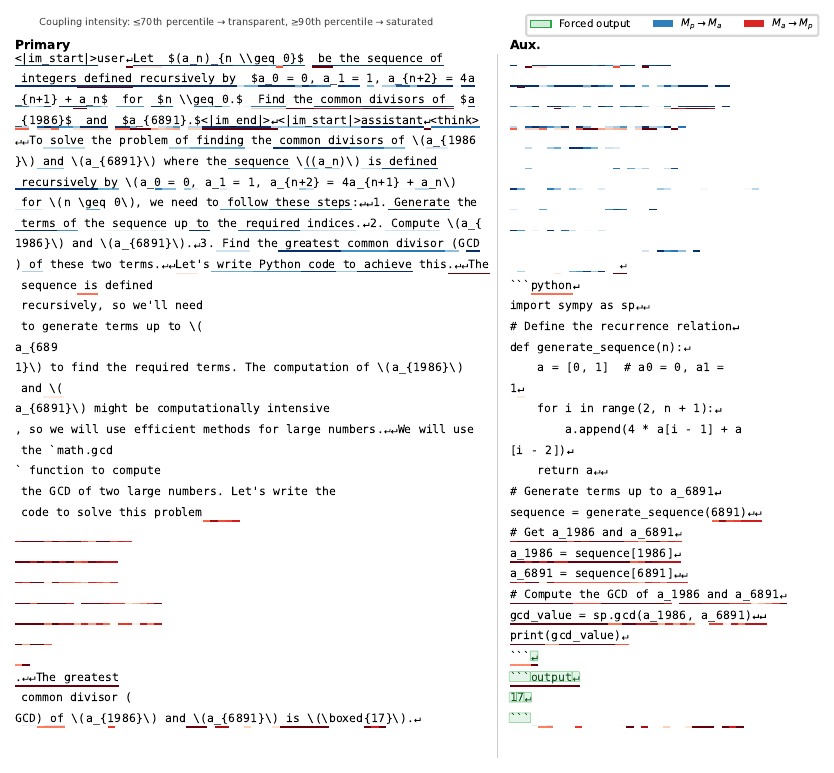}
\caption{%
\textbf{Full coupling trace: GCD of recurrence sequence terms} (twin Qwen3-4B, identity interface, Python tool).
The auxiliary (right) waits with spaces while the primary processes the problem, then emits a correct Python program that generates the sequence and computes $\gcd(a_{1986}, a_{6891}) = 17$.
The recurrence relation, initial conditions, and both target indices were recovered entirely from hidden-state signals; the auxiliary never sees the problem text.
Forward coupling (blue) is elevated during the primary's problem-processing phase. Reverse coupling (red) spikes in two contexts: while the auxiliary is still writing code, guiding the primary to emit space tokens and wait for completion; and when the primary model produces its boxed answer, at which point the result flows over from the auxiliary.
Row breaks in the visualization are dictated by the auxiliary's explicit newlines for code readability; additional wrapping within rows is due to horizontal space constraints maintaining token-level synchronization.
}
\label{fig:full_python_gcd}
\end{figure*}

\begin{figure*}[h]
\centering
\includegraphics[width=\linewidth]{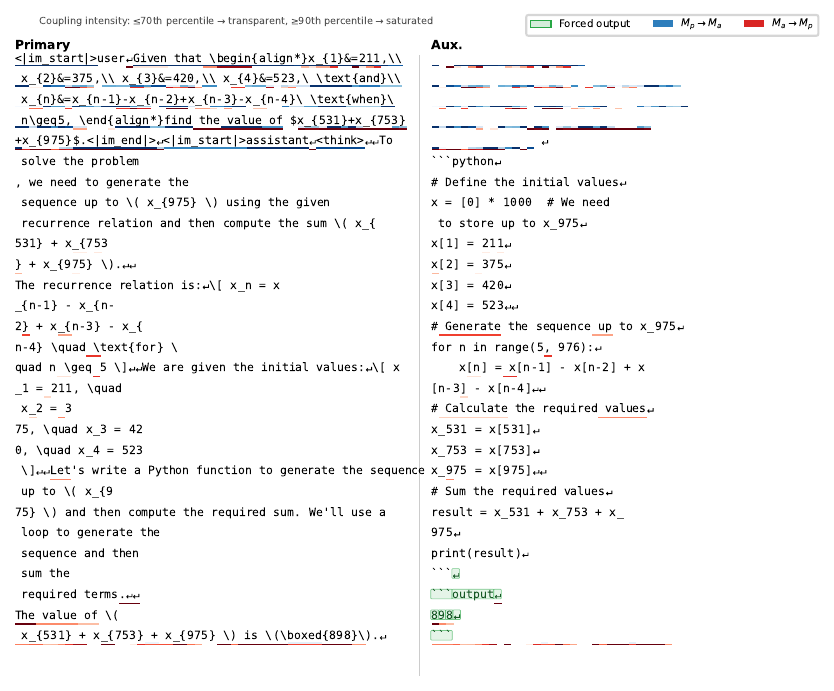}
\caption{%
\textbf{Full coupling trace: recursive sequence evaluation} (twin Qwen3-4B, identity interface, Python tool).
The auxiliary (right) recovers all seven parameters (4 initial values and 3 target indices) plus the recurrence structure from hidden states alone, generating correct Python code that yields 898.
This example demonstrates high-bandwidth information transfer through the neural channel: 7 distinct numerical values and one structural relation are communicated without any text exchange.
}
\label{fig:full_python_recurrence}
\end{figure*}

\begin{figure*}[h]
\centering
\includegraphics[width=\linewidth]{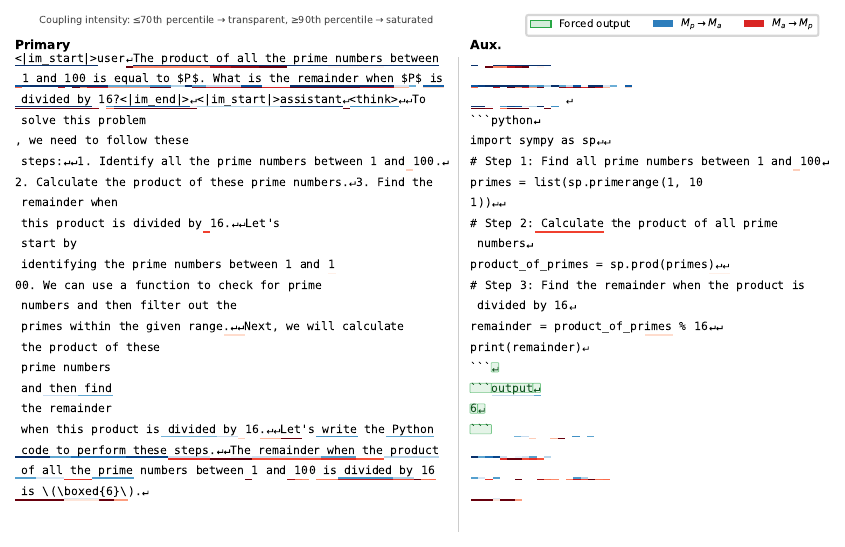}
\caption{%
\textbf{Full coupling trace: prime product modular arithmetic} (twin Qwen3-4B, identity interface, Python tool).
The auxiliary (right) generates a concise program using \texttt{sympy.primerange} and modular reduction, producing the correct answer of 6.
Unlike the previous two examples which require recovering many numerical parameters, this problem tests whether the channel can transmit the \emph{structural} content of a number-theoretic question (the concept of ``product of primes in a range, reduced modulo 16'') without any text.
}
\label{fig:full_python_remainder}
\end{figure*}

Figures~\ref{fig:full_python_gcd}--\ref{fig:full_python_remainder} show complete traces for the Python-tool configuration (twin Qwen3-4B), where the auxiliary model receives \emph{no problem text}, only hidden-state signals at intermediate layers. Despite this, it generates problem-specific Python code.

\FloatBarrier

\end{document}